\documentclass[acmtog,nonacm]{acmart}

\usepackage[normalem]{ulem}
\usepackage{xcolor}
\usepackage{siunitx}
\usepackage{booktabs}
\usepackage{multirow}
\usepackage{comment}
\usepackage{array}
\usepackage{pifont}
\usepackage{float}
\usepackage{xspace}
\newcommand{\cmark}{\ding{51}}
\newcommand{\xmark}{\ding{55}}
\newcommand{\method}{\textsc{RecGen3D}\xspace} 

\AtBeginDocument{%
  }

\acmJournal{TOG}

\setcopyright{none}
\begin{document}

\title{RecGen3D: Reconstruction-Guided 3D Generation in a Shared Canonical Space}

\author{Zhisheng Huang}
\orcid{0000-0001-5791-5621}
\email{hzs@tamu.edu}
\affiliation{\institution{Texas A\&M University}\city{College Station}\country{USA}}

\author{Jiahao Chen}
\orcid{0009-0000-4928-8285}
\email{jiahao.chen@tamu.edu}
\affiliation{\institution{Texas A\&M University}\city{College Station}\country{USA}}

\author{Cheng Lin}
\orcid{0000-0002-3335-6623}
\email{chenglin@must.edu.mo}
\affiliation{\institution{Macau University of Science and Technology}\city{Macau}\country{Macau}}

\author{Chenyu Hu}
\orcid{0009-0004-7636-4174}
\email{hcy299792@gmail.com}
\affiliation{\institution{HKUST}\city{Hong Kong}\country{Hong Kong}}

\author{Hanzhuo Huang}
\orcid{0009-0005-7175-9342}
\email{huanghzh2022@shanghaitech.edu.cn}
\affiliation{\institution{ShanghaiTech University}\city{Shanghai}\country{China}}

\author{Zhengming Yu}
\orcid{0009-0003-0553-8125}
\email{yuzhengming@tamu.edu}
\affiliation{\institution{Texas A\&M University}\city{College Station}\country{USA}}

\author{Mengfei Li}
\orcid{0009-0008-8852-8999}
\email{mliek@connect.ust.hk}
\affiliation{\institution{HKUST}\city{Hong Kong}\country{Hong Kong}}

\author{Yuheng Liu}
\orcid{0009-0001-3690-5659}
\email{yuhenl22@uci.edu}
\affiliation{\institution{University of California Irvine}\city{Irvine}\country{USA}}

\author{Zekai Gu}
\orcid{0000-0002-9739-1209}
\email{zekai.gu@u.nus.edu}
\affiliation{\institution{HKUST}\city{Hong Kong}\country{Hong Kong}}

\author{Zibo Zhao}
\orcid{0009-0005-4831-8418}
\email{zhaozb@shanghaitech.edu.cn}
\affiliation{\institution{ShanghaiTech University}\city{Shanghai}\country{China}}

\author{Yuan Liu}
\orcid{0000-0003-2933-5667}
\email{yuanly@ust.hk}
\affiliation{\institution{HKUST}\city{Hong Kong}\country{Hong Kong}}

\author{Xin Li}
\orcid{0000-0002-0144-9489}
\email{xinli@tamu.edu}
\author{Wenping Wang}
\authornote{Corresponding author.}
\orcid{0000-0002-2284-3952}
\email{wenping@tamu.edu}
\affiliation{\institution{Texas A\&M University}\city{College Station}\country{USA}}

\begin{abstract}
Sparse-view 3D modeling represents a fundamental tension between reconstruction fidelity and generative plausibility. While feed-forward reconstruction excels in efficiency and input alignment, it often lacks the global priors needed for structural completeness. Conversely, diffusion-based generation provides rich geometric details but struggles with multi-view consistency.
We present RecGen3D, a framework that combines these two paradigms into a cooperative system. To overcome inherent conflicts in coordinate spaces, 3D representations, and training objectives, we align both models within a shared canonical space. We employ decoupled cooperative learning, which maintains stable training while enabling seamless collaboration during inference. Specifically, the reconstruction module is adapted to provide canonical geometric anchors, while the diffusion generator leverages latent-augmented conditioning to refine and complete the geometric structure.
Experimental results demonstrate that RecGen3D achieves superior fidelity and robustness, outperforming existing methods in creating complete and consistent 3D models from sparse observations.
\end{abstract}

\vspace{-0.5cm}


\begin{teaserfigure}
  \vspace{-0.2cm}
  \centering
  \includegraphics[width=0.88\textwidth]{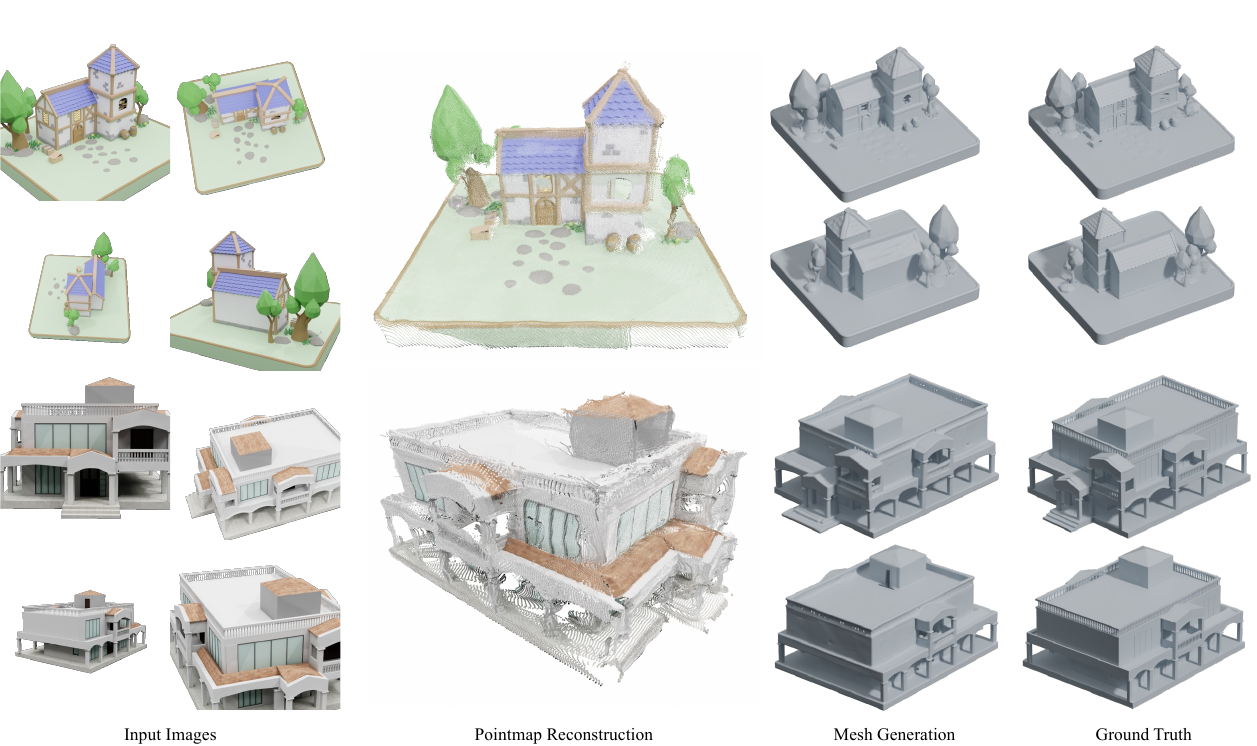}
  \vspace{-0.1cm}
  \caption{\textbf{RecGen3D enables high-fidelity 3D object reconstruction from sparse, unposed images.} It first establishes a deterministic point-cloud ``geometric anchor'' within a canonical coordinate system. Guided by this anchor, a generative model synthesizes a detailed mesh that preserves instance-specific structural features while producing plausible geometric completions for unobserved regions, ensuring high alignment with the ground-truth shape.}
  \label{fig:teaser}
\end{teaserfigure}

\maketitle
\renewcommand{\shortauthors}{Huang et al.}

\section{Introduction}
\label{sec:intro}

The quest for high-fidelity 3D object modeling from sparse observations has long been a cornerstone of computer graphics and computer vision. This pursuit has branched into two distinct paradigms: deterministic reconstruction and stochastic generation. The former, typified by recent feed-forward regression models \cite{wang2025vggt}, focuses on fidelity by lifting 2D observations into 3D space. However, these methods often suffer from geometric incompleteness and lack the semantic consistency due to their reliance on local features and lack of global structural priors. In contrast, the emergence of 3D diffusion models has introduced powerful object-centric priors, enabling the synthesis of plausible and detailed geometries. Yet, these generative approaches often struggle to remain strictly faithful to the specific fine-grained details provided in the input images, leading to ``hallucinations'' that deviate from the actual object geometry.

Recently, a prominent trend has emerged seeking to unify these paradigms by aligning the latent spaces of 2D generative models with 3D reconstruction foundations. Notable works such as VIST3A~\cite{go2025vist3a} and Gen3R~\cite{huang2026gen3r} employ model stitching or adapter-based regularization to ensure that the outputs of 2D video diffusion models are inherently decodable by 3D reconstruction systems. While these methods successfully make 2D generation ``3D-aware'', they remain fundamentally tethered to the 2D domain. Consequently, they often inherit the ``representation gap'' and consistency artifacts typical of 2D-to-3D lifting.
In this paper, we propose a departure from alignment-based lifting in favor of a native-3D \emph{reconstruct-then-condition} integration: we make a feed-forward geometric reconstructor and a native 3D generator compatible within a shared canonical space, so that explicit reconstructed geometry directly steers generation. We argue that 3D reconstruction and 3D generation are not competing methodologies, but inherently complementary dimensions of the same 3D inference problem.

Despite the conceptual elegance of this integration, achieving {synergy} between these two systems presents two fundamental technical challenges: The first challenge is \emph{Disparate Learning Dynamics}. Reconstruction models rely on deterministic regression, while diffusion models operate through stochastic iterative denoising. A naive joint training of these different objectives  leads to slow convergence and prohibitive computational overhead, as each module must adapt to the fluctuating and noisy outputs of the other.
The second challenge is \emph{Representation and Coordinate Incompatibility}: Feed-forward reconstruction typically operates in camera-relative spaces using explicit geometric features, whereas diffusion-based generation is usually performed in a compact latent space
{where the latent features are mapped to and interpreted within a canonical, object-centric coordinate system}. This misalignment prevents effective information flow and structural collaboration between the two modules.

To address disparate learning dynamics, we adopt a decoupled modular design for a holistic 3D system. In this framework, the reconstruction model is trained first to provide a stable geometric anchor that preserves fidelity to the input observations. The generation model is then trained as a prior-driven refiner to recover unobserved regions and synthesize fine-grained details.

As a result, the outputs remain faithful to the evidence while achieving complete, high-fidelity synthesis. Beyond performance, the modular design offers long-term utility: sub-modules can be swapped as stronger models emerge,  and external priors can be injected without retraining the entire system.
This decoupling is deliberate: joint training under identical backbones, data, and budget---even from the same pretrained weights---degrades every geometric metric (supplementary Sec.~S8).

To resolve the second challenge of representation and coordinate incompatibility, we establish a shared canonical 3D modeling space that serves as a shared structural bridge. The difficulty lies not in placing two strong models side by side, but in bridging their incompatibility: camera-relative explicit geometry versus canonical-space latent representations. Through proposed branch repurposing strategy, we transform the reconstruction model's native coordinate system into this canonical space, ensuring its outputs are natively interpretable by the generative framework.
This shared foundation enables a latent-augmented multi-view conditioning strategy, where the reconstruction results—specifically pointmaps and intermediate features—serve as a robust geometric scaffold to guide the 3D diffusion process. The key insight behind this design is that while point-based signals provide essential constraints for 3D spatial reasoning, they are discontinuous or noisy in capturing fine-grained details. We therefore incorporate multi-view DINO tokens to maintain the high-fidelity synthesis capabilities of the base model. To ensure spatial consistency across these views, geometric tokens from the reconstruction model act as structural anchors.
Unlike the prior reconstruct-then-condition method ReconViaGen~\cite{chang2025reconviagen}, which injects \emph{implicit} reconstruction features, our generator is conditioned on \emph{explicit} canonical geometry.

Our main contributions are summarized as follows:

\begin{itemize}
  \item We introduce RecGen3D, a reconstruct-then-condition framework that couples feed-forward 3D reconstruction with native 3D generation in a shared canonical space for object shape modeling from unposed multi-view images.
  \item We propose to adopt a decoupled modular design to avoid disparate learning dynamics during the integration, validated by a controlled joint-training comparison.
  \item We establish a shared canonical 3D space by leveraging the branch repurposing and the latent-augmented multi-view conditioning strategy to address representation and coordinate incompatibility during integration.
  \item Experimental results demonstrate that our methods significantly outperform SOTA methods in both multi-view consistency and mesh generation quality.
\end{itemize}

\noindent Code and pretrained weights for this paper are at \url{https://github.com/zsh523/RecGen3D}.

\vspace{-0.2cm}
\section{Related Work}

\subsection{3D Generation}

The field of 3D content generation has rapidly shifted from costly per-scene optimization to efficient, scalable generative frameworks capable of producing high-quality assets in seconds.
Initial efforts leveraged Generative Adversarial Networks (GANs)~\cite{goodfellow2014generative} to learn distributions over various 3D representations~\cite{wu2016learning, chan2022efficient, gao2022get3d}, but these struggled to scale to diverse and complex shapes.
The introduction of diffusion models~\cite{ho2020denoising, sohl2015deep} marked a turning point, enabling higher-quality generation across multiple 3D representations including point clouds~\cite{nichol2022point, luo2021diffusion, vahdat2022lion}, voxel grids~\cite{hui2022neural, muller2023diffrf}, neural radiance fields~\cite{mildenhall2020nerf, chen2023single}, and signed distance functions~\cite{park2019deepsdf, zheng2022sdf}. More recently, 3D Gaussian Splatting (3DGS)~\cite{kerbl20233d} has emerged as an efficient representation inspiring new generative models~\cite{zhang2024gaussiancube, he2024gvgen, tang2024lgm}.

To reduce the cost of high-dimensional 3D generation, recent methods adopt latent diffusion approaches~\cite{rombach2022high} that compresses 3D data into compact latent spaces for efficient modeling.
Current native 3D generative methods broadly fall into three main categories: vector set-based methods~\cite{zhang20233dshape2vecset,zhang2024clay,zhao2024michelangelo,wu2024direct3d,li2024craftsman,lan2024ln3diff,DBLP:journals/corr/abs-2403-02234, li2025triposg, chen2025meshgengeneratingpbrtextured, zhao2025hunyuan3d20scalingdiffusion,hunyuan3d2025hunyuan3d, lai2025hunyuan3d25highfidelity3d,yang2025holopartgenerative3damodal,lin2025partcrafterstructured3dmesh,tang2025efficientpartlevel3dobject}, sparse voxel-based methods~\cite{xiang2025structured3dlatentsscalable,ye2025hi3dgen,he2025sparseflexhighresolutionarbitrarytopology3d, wu2025direct3ds2gigascale3dgeneration, li2025sparc3dsparserepresentationconstruction}, and autoregressive mesh generation~\cite{siddiqui2024meshgpt,chen2024meshanythingartistcreatedmeshgeneration,chen2024meshxl,tang2024edgerunnerautoregressiveautoencoderartistic,chen2024meshanythingv2artistcreatedmesh,weng2024scalingmeshgenerationcompressive,hao2024meshtronhighfidelityartistlike3d,wang2024llamameshunifying3dmesh,zhao2025deepmeshautoregressiveartistmeshcreation}.
Most recently, native 3D generation has advanced toward compact structured latents~\cite{xiang2026native} and deployable systems that generate assets at interactive speed~\cite{wang2026assetgen, feng2026fastsam3d}.

An alternative paradigm leverages pretrained 2D diffusion models via score distillation sampling~\cite{poole2023dreamfusion, wu2024reconfusion, wang2023prolificdreamer}, but it relies on costly test-time optimization and often yields inconsistent geometry~\cite{wang2023prolificdreamer, tang2023make}.
Multi-view diffusion models~\cite{li2023instant3d, liu2023one, xu2024instantmesh, shi2024mvdream, long2024wonder3d, liu2023zero, voleti2024sv3d, chen2024v3d,yang2025prometheus, schwarz2025generative, liang2025wonderland, lin2025diffsplat, szymanowicz2025bolt3d} generate consistent views for 3D reconstruction but typically produce less accurate geometry than native 3D methods.

Despite rapid progress in 3D generation, recovering high-fidelity 3D assets from multiple unposed RGB images remains underexplored, as it requires fine geometry and strong multi-view consistency without known camera poses.
Existing combinations of the two paradigms fall into two broad families: \emph{generate-then-reconstruct} methods synthesize observations and reconstruct from them~\cite{li2023instant3d, xu2024instantmesh, long2024wonder3d, voleti2024sv3d}, with VIST3A~\cite{go2025vist3a} and Gen3R~\cite{huang2026gen3r} as latent-alignment variants, whereas \emph{reconstruct-then-condition} methods extract geometric evidence to steer a native 3D generator, via explicit control interfaces (point-cloud conditioning in Hunyuan3D-Omni~\cite{hunyuan3d2025hunyuan3domni}) or implicit feature injection (ReconViaGen~\cite{chang2025reconviagen}, which feeds multi-level features from a fine-tuned VGGT into a TRELLIS-based generator). Unified photogrammetry models such as Matrix3D~\cite{lu2025matrix3d} cast reconstruction subtasks generatively within one diffusion backbone. Our method belongs to the second family: unlike pose-free PF-LRM~\cite{wang2024pf}, constrained by a Triplane NeRF representation, and ReconViaGen's multi-stage implicit design, RecGen3D canonicalizes VGGT predictions and conditions Hunyuan3D-Omni directly on explicit canonical geometry with latent-augmented multi-view features---a simpler and more interpretable pipeline for unposed inputs.

\subsection{Multi-View 3D Reconstruction}

Traditional 3D reconstruction typically relies on Structure-from-Motion~\cite{schonberger2016structure} and Multi-View Stereo~\cite{schoenberger2016mvs} to recover geometry via photometric consistency and triangulation. While learning-based MVS methods~\cite{yao2018mvsnet, yao2019recurrent, gu2020cascade, wang2021patchmatchnet} have improved efficiency through deep cost volume regularization, they remain sensitive to camera pose accuracy and often struggle with sparse inputs~\cite{tanNOPESACNeuralOnePlane2023}. Alternatively, Neural Radiance Fields~\cite{mildenhall2020nerf} enable high-quality synthesis via differentiable rendering, yet they frequently suffer from overfitting under sparse views and require costly per-scene optimization despite various regularization strategies~\cite{niemeyerRegNeRFRegularizingNeural2022, yu2021pixelnerf, wangIBRNetLearningMultiView2021}.

The advent of 3DGS~\cite{kerbl20233d} has inspired feed-forward architectures that directly regress Gaussian parameters from sparse multi-view features~\cite{charatan2024pixelsplat, tangHiSplatHierarchical3D2024, zhangTranSplatGeneralizable3D, jenaSparSplatFastMultiView2025}, achieving impressive speed and generalization. However, these approaches fundamentally rely on local geometric features and lack strong cross-view priors. Consequently, they often produce incomplete reconstructions with visible artifacts in occluded regions, particularly when view overlap is minimal.

A significant bottleneck remains the dependency on external pose estimation. While some pose-free methods attempt to jointly optimize camera parameters and geometry~\cite{lin2021barf, fanInstantSplatUnboundedSparseview, fuCOLMAPFree3DGaussian}, they are often prone to local minima. Recent stereo foundation models, such as DUSt3R~\cite{wang2024dust3r}, MASt3R~\cite{leroy2024grounding}, and VGGT~\cite{wang2025vggt}, address this by predicting aligned point maps via dense correspondence. This family continues to advance rapidly, with permutation equivariance ($\pi^3$~\cite{wang2025pi}), dynamic scenes and streamlined depth-ray targets (D4RT~\cite{zhang2025d4rt}, Depth Anything~3~\cite{lin2025depthanything3}), flexible geometric inputs (MapAnything~\cite{keetha2026mapanything}), and predictable scaling (VGGT-$\Omega$~\cite{wang2026vggtomega}). Although these models enable robust pose-free reconstruction, they typically yield point clouds that lack surface topology or completeness.

In contrast to methods relying solely on per-view geometry, our work integrates feed-forward reconstruction with generative 3D priors learned from large-scale data.  Our framework is agnostic to the reconstruction backbone: we instantiate it with VGGT and show that the same recipe transfers to $\pi^3$ (supplementary Sec.~S9).
\section{Method}

\begin{figure*}[t]
    \centering
    \vspace{-0.4cm}
    \includegraphics[width=\textwidth]{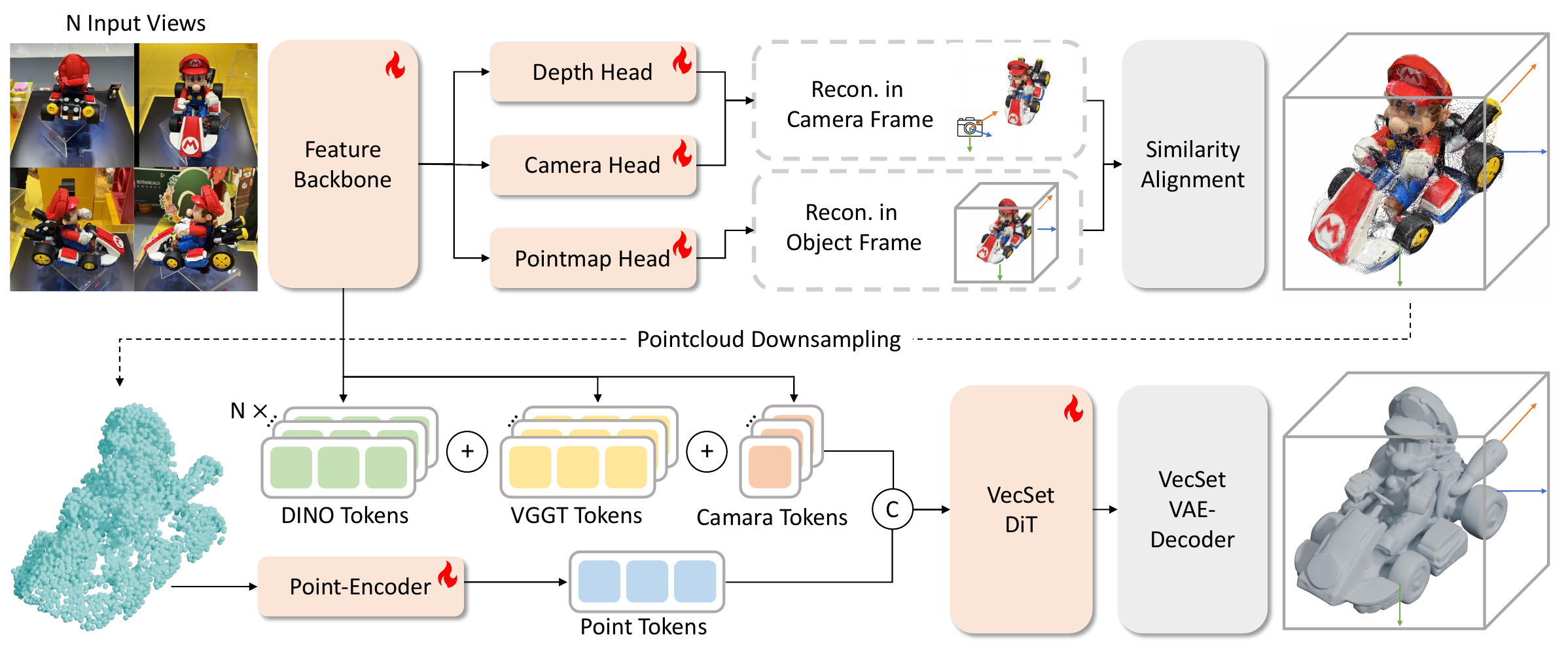}
    \vspace{-0.8cm}
    \caption{\textbf{Overview of our method.} Given $N$ unposed input views, we first canonicalize feed-forward multi-view geometry predictions via branch repurposing and similarity alignment to obtain a canonical point cloud (top). We then train a controllable 3D generator conditioned on this point cloud together with multi-view image features, geometry latents, and camera embeddings to synthesize a high-fidelity mesh (bottom).}
    \label{fig:pipe}
    \vspace{-0.4cm}
\end{figure*}

Given an arbitrary number of unposed input images of an object, our method aims to recover high-fidelity 3D shape by coupling feed-forward multi-view geometry estimation with controllable 3D generation in a reconstruct-then-condition pipeline.
A fundamental challenge is that feed-forward geometry foundations typically predict camera-centric, view-dependent outputs that are not directly compatible with object-centric generative priors; meanwhile, 3D generators assume canonical object space and require reliable geometric guidance to remain faithful to the input views.
To bridge this gap, we propose a two-stage modular pipeline.
First, we canonicalize the feed-forward geometry predictions via a branch-repurposing strategy, producing {generation-compatible} canonical point maps and a refined canonical point cloud through similarity alignment (Sec.~\ref{sec:method_recon}).
Second, we train a controllable 3D generator with {latent-augmented multi-view conditioning}, combining dense image tokens with geometry and camera embeddings to inject multi-view geometric context while retaining strong appearance priors (Sec.~\ref{sec:method_gen}).
This modular design preserves pretrained priors, supports model interchangeability, and enables effective geometry-guided generation.
An overview of our method is presented in Fig.~\ref{fig:pipe}.

\subsection{Preliminaries}
\label{pre}

\vspace{1mm}
\noindent\textbf{Feed-Forward 3D Reconstruction.} 
Feed-forward geometry models such as {VGGT}~\cite{wang2025vggt}, Fast3R~\cite{yang2025fast3r}, and $\pi^{3}$~\cite{wang2025pi} leverage large-scale Transformers to jointly regress camera parameters and scene geometry across an arbitrary number of views in a single pass. We adopt this scalable feedforward multiview representation as our foundation, choosing {VGGT} as a representative instantiation due to its inherent efficiency and ability to maintain global geometric consistency without iterative optimization.
Given multi-view images $\left\{\mathbf{I}_i\in \mathbb{R}^{H \times W \times 3}\right\}_{i=1}^{N}$, VGGT first extracts their spatial features $\mathbf{F}_{i}^{D} \in \mathbb{R}^{K\times C}$ via DINO~\cite{oquab2023dinov2} and concatenates them with additional camera tokens $\mathbf{t}_i^{\text{cam}} \in \mathbb{R}^{1\times C}$, where $K$ is the spatial token number for each image and $C$ is the feature dimension.
These tokens are later processed through a large transformer to obtain VGGT tokens $\mathbf{F}_{i}^{V} \in \mathbb{R}^{K\times C}$ and refined camera tokens $\hat{\mathbf{t}}_i^{\text{cam}} \in \mathbb{R}^{1\times C}$.
For different tasks, VGGT adopts three prediction heads:
(1) the \textit{Camera head} that maps camera tokens $\hat{\mathbf{t}}_i^{\text{cam}}$ to camera parameters $\mathbf{g}_{i}^{\text{ref}} \in \mathbb{R}^{9}$ in the first camera's reference frame;
(2) the \textit{Depth head} that maps VGGT tokens $\mathbf{F}_{i}^{V}$ to pixel-wise depth maps $\mathbf{D}_{i}^{\text{ref}} \in \mathbb{R}^{H\times W}$;
and (3) the \textit{Point head} that maps VGGT tokens $\mathbf{F}_{i}^{V}$ to viewpoint-invariant point maps $\mathbf{P}_{i}^{\text{ref}} \in \mathbb{R}^{H \times W \times 3}$ in the reference frame.
The entire model is optimized end-to-end using multi-task regression losses:
\begin{equation}
    \mathcal{L} = \mathcal{L}_\text{camera} + \mathcal{L}_\text{depth} + \mathcal{L}_\text{pmap} + \lambda \mathcal{L}_\text{track}.
\end{equation}

\vspace{1mm}
\noindent\textbf{Controllable 3D Generation.}
 While our framework is designed to be compatible with various 3D generative backends, we utilize Hunyuan3D-Omni~\cite{hunyuan3d2025hunyuan3domni} as a representative instantiation. By leveraging its multi-modal control encoder, we can directly incorporate geometric guidance while bypassing the substantial overhead of training a conditioned generative prior from scratch.
Hunyuan3D-Omni operates by first mapping 3D geometry into a structured latent space, followed by a diffusion process to synthesize novel shapes.
Given a surface-sampled point set $\mathbf{S}\in\mathbb{R}^{N\times 6}$ that concatenates 3D coordinates with surface normals, the Transformer-based encoder maps $\mathbf{S}$ into a sequence of latent tokens $\mathbf{Z}\in\mathbb{R}^{L\times C}$.
These latent tokens are subsequently processed by a diffusion transformer under a multi-modal conditioning signal $cond = [\mathbf{F}^{D}, \beta]$, where $\mathbf{F}^{D}$ represents DINO-extracted image features and $\beta$ denotes the embedding produced by the unified control encoder for point-cloud, voxel, bounding-box, and skeletal conditions.
To produce the final output, the denoised representation is decoded into a signed distance field $F_{\mathrm{sdf}}$, from which high-quality triangular meshes are extracted via marching cubes~\cite{lorensen1987marching}.
The framework is optimized end-to-end using a flow-matching objective:
\begin{equation}
    \mathcal{L} = \mathbb{E}_{t, x_0, x_1, cond} \left\Vert u_\theta (x_t, t, cond) - (x_1 - x_0) \right\Vert_{2}^{2},
    \label{eq:diffusion}
\end{equation}
where $x_0$ is Gaussian noise, $x_1$ is the latent sample, and $x_t = (1-t)x_0 + tx_1$ describes the trajectory under $t \sim \mathcal{U}(0,1)$.

\subsection{Generation-compatible Feed-Forward Reconstruction}
\label{sec:method_recon}

Feed-forward 3D reconstruction methods\cite{wang2025pi, wang2025vggt,yang2025fast3r} demonstrate robust open-world generalization by leveraging massive, diverse training corpora.
A shared characteristic is their multi-head prediction architecture. In the case of \textsc{VGGT}, this involves a combination of relative camera poses ($\mathbf{g}_i^{\text{ref}}$), per-view depth maps ($\mathbf{D}_i^{\text{ref}}$), and global point maps ($\mathbf{P}_i^{\text{ref}}$).
To maintain scene-agnostic flexibility, these architectures typically represent geometry within a relative reference frame, often defined by the coordinate system of the primary input camera. Conversely, 3D generative frameworks model shapes in a canonical, object-centered space to ensure consistent orientation, scale, and topological integrity.

We address this mismatch by canonicalizing the feed-forward reconstruction prediction.

\vspace{1mm}
\noindent \textbf{Design Space Exploration.} We explore three approaches to achieve canonical space prediction:
\begin{itemize}
    \item \textit{Direct supervision transferring}: Directly supervise all heads in canonical space, but this causes degradation of pretrained 3D priors.
    \item \textit{Explicit transformation prediction}: Add dedicated tokens and heads to predict reference-to-canonical transformation, but in our setting weak gradient signals lead to slow convergence and poor performance.
    \item \textit{Branch repurposing (ours)}: Repurpose only the point map head to output $\mathbf{P}_i^{\text{can}}$ in canonical space, while keeping $\mathbf{g}_i^{\text{ref}}$ and $\mathbf{D}_i^{\text{ref}}$ in the reference frame.
\end{itemize}

\vspace{1mm}
\noindent \textbf{Our Approach.} We adopt branch repurposing based on three key insights:
(1) Maintaining $\mathbf{g}_i^{\text{ref}}$ and $\mathbf{D}_i^{\text{ref}}$ in the reference frame provides strong regularization preserving VGGT's learned priors.
(2) Supervising only $\mathbf{P}_i^{\text{can}}$ in canonical space provides dense gradient signals for effective adaptation---a dense-target principle shared with Depth Anything~3~\cite{lin2025depthanything3}.
(3) We can transform the more accurate depth predictions to canonical space via similarity alignment with $\mathbf{P}_i^{\text{can}}$, obtaining better geometry than using $\mathbf{P}_i^{\text{can}}$ alone.
The weakness of explicit transformation prediction does not contradict D4RT~\cite{zhang2025d4rt} or Depth Anything~3~\cite{lin2025depthanything3}: their transforms are pinned down by multi-view correspondence, whereas the reference-to-canonical one is semantics-dependent and ill-posed under symmetry.
Specifically, we align depth-derived 3D points (from $\mathbf{g}_i^{\text{ref}}$ and $\mathbf{D}_i^{\text{ref}}$) with predicted canonical points $\mathbf{P}_i^{\text{can}}$.
We apply two-stage sampling: uniform sampling yields $4M$ points, then Farthest Point Sampling (FPS) selects $M$ spatially diverse correspondences. 

The similarity transformation $\mathbf{T} \in \text{Sim}(3)$ is computed via weighted Procrustes analysis:

\begin{equation} 
\mathbf{T} = \mathop{\text{argmin}}_{\mathbf{T} \in \text{Sim}(3)} \sum_{j=1}^{M} w_j \|\mathbf{P}_j^{\text{can}} - \mathbf{T} (\mathbf{D}_{j}^{\text{ref}} )\|^2,
\end{equation}
where $w_i$ denotes confidence weights, $\mathbf{D}_j^{\text{ref}}$ represents the $j$-th depth-derived 3D point, and $\text{Sim}(3)$ encompasses rotation, translation, and uniform scaling.

\subsection{Reconstruction-guided Controllable Generation}
\label{sec:method_gen}

Having obtained a canonicalized 3D reconstruction from a feed-forward model, we next leverage these results as conditioning signals for controllable 3D generation. We adopt Hunyuan3D-Omni (see Section~\ref{pre}), which supports point-cloud conditioning to guide shape synthesis. This design naturally allows us to use the generation-compatible pointmap as direct geometric control.
However, without further training, we observe that the generation model fails to deliver high-fidelity 3D object generation due to the following:
(1) VGGT 3D reconstruction features outliers and noise, this kind of point cloud conditioning is not seen by the Hunyuan3D-Omni generation model during training.
(2) Point cloud conditioning could only provide limited geometry details as it doesn't contain continuous surface information.
(3) Hunyuan3D-Omni is limited to single view conditioning, failing to leverage multi-view image information.

To this end, we further train the Hunyuan3D-Omni generation model to fully leverage the 3D object reconstruction results. 
Specifically, we modify the model to condition on both point-cloud and multi-view images. We keep the $\beta$ (See Section~\ref{pre}) unchanged, 
while modifying the joint conditioning to
\begin{equation}
\label{eq:joint_condition_modified}
cond = [\mathbf{F}_1^{MV}, \mathbf{F}_2^{MV}, \ldots, \mathbf{F}_N^{MV}, \beta],
\end{equation}
where $\mathbf{F}_i^{MV}$ denotes the multi-view conditioned features for the $i$-th frame.
This simple yet effective approach allows us to fully exploit the strengths of the base Hunyuan3D-Omni model by retaining the original point cloud conditioning pathway.
The DiT attends to this concatenated sequence through its native cross-attention, requiring no architectural changes; the exact token layout and encoder specifications are given in supplementary Sec.~S11.

\vspace{1mm}
\noindent \textbf{Design Space Exploration.} We investigate two distinct paradigms for integrating multi-view geometric conditioning into the generative pipeline:

\begin{itemize}
    \item \textit{Point-Guided Feature Sampling}: In this configuration, the conditioning points $\mathbf{P}^c$ (downsampled from the initial VGGT reconstruction) are used to index a corresponding subset of multi-view DINO tokens. This aims to strike a balance between image-derived semantic signals and explicit geometric constraints. These features are further augmented with 3D positional encodings of $\mathbf{P}_i^c$ to provide spatial grounding:
        \begin{equation}
        \label{eq:subsampled_features}
        \mathbf{F}_i^{MV} = \mathcal{S}(\mathbf{F}_i^D, \mathbf{P}^c) + \text{MLP}(\text{PE}(\mathbf{P}^c)),
        \end{equation}
        where $\mathcal{S}(\cdot, \mathbf{P}^c)$ denotes the spatial sampling operator indexed by points $\mathbf{P}^c$.

    \item \textit{Latent-Augmented View Conditioning (Ours)}: Rather than sparsifying the feature space, we enrich the full set of DINO tokens $\mathbf{F}_i^D$ by injecting view-dependent geometric context. We project VGGT latent tokens and camera tokens via learnable MLPs to serve as geometric embeddings:
        \begin{equation}
        \label{eq:vggt_embedded_features}
        \mathbf{F}_i^{MV} = \mathbf{F}_i^D + \text{MLP}_{\text{view}}(\mathbf{F}_i^V) + \text{MLP}_{\text{cam}}(\hat{\mathbf{t}}_i^{\text{cam}}). 
        \end{equation}
\end{itemize}

\vspace{1mm}
\noindent \textbf{Our Approach.} We adopt the latent-augmented approach based on two key insights:
(1) Hunyuan3D-Omni's synthesis quality relies on dense semantic priors within DINO tokens rather than sparse geometric constraints, which stems from the fact that Hunyuan3D-Omni is built upon Hunyuan3D 2.1, an image-to-3D generation model. Besides, images inherently contain richer structural information than point clouds.
(2) VGGT tokens $\mathbf{F}_i^V$ represent a continuous, learned geometric manifold with global structural coherence and multi-scale spatial context, which enables the diffusion model to better integrate multi-view information.

\begin{table*}[t]
\centering
\caption{\textbf{Quantitative Results for 3D Object Reconstruction.} We report geometric metrics across the Toys4K and GSO datasets. Our framework consistently outperforms existing generative and reconstruction-based methods across all metrics.}
\label{tab:main_results}
\vspace{-2mm}
\small
\begin{tabular}{llcccccc}
\toprule
Dataset & Method & Chamfer-$L_2 \downarrow$ & Prec. $\uparrow$ & Rec. $\uparrow$ & F-Score $\uparrow$ & Normal $\uparrow$ & IoU $\uparrow$ \\
\midrule
\multirow{7}{*}{Toys4K} 
& LucidFusion \cite{he2024lucidfusion}    & 0.1333 & 0.2143 & 0.1378 & 0.1471 & 0.6240 & 0.1035 \\
& Hunyuan3D-MV \cite{zhao2025hunyuan3d20scalingdiffusion}    & 0.0513 & 0.4930 & 0.4305 & 0.4540 & 0.8124 & 0.6311 \\
& TRELLIS-M \cite{xiang2025structured3dlatentsscalable}     & 0.0343 & 0.5952 & 0.5600 & 0.5674 & 0.8309 & 0.6318 \\
& TRELLIS-S  \cite{xiang2025structured3dlatentsscalable}    & 0.0309 & 0.5762 & 0.5430 & 0.5525 & 0.8324 & 0.6583 \\
& SAM 3D  \cite{chen2025sam}   & 0.0354 & 0.5416 & \underline{0.6575} & 0.5711 & 0.8218 & 0.6061 \\
& ReconViaGen \cite{chang2025reconviagen}   & \underline{0.0281} & \underline{0.6224} & 0.6086 & \underline{0.6105} & \underline{0.8543} & \underline{0.7229} \\
& \textbf{Ours}  & \textbf{0.0175} & \textbf{0.7622} & \textbf{0.7834} & \textbf{0.7695} & \textbf{0.8785} & \textbf{0.8030} \\
\midrule
\multirow{7}{*}{GSO} 
& LucidFusion \cite{he2024lucidfusion}   & 0.1203 & 0.2004 & 0.1648 & 0.1650 & 0.6421 & 0.1317 \\
& Hunyuan3D-MV \cite{zhao2025hunyuan3d20scalingdiffusion}    & 0.0667 & 0.4060 & 0.3680 & 0.3781 & 0.7897 & 0.5980 \\
& TRELLIS-M  \cite{xiang2025structured3dlatentsscalable}    & 0.0318 & 0.5071 & 0.4863 & 0.4918 & 0.8482 & 0.6489 \\
& TRELLIS-S  \cite{xiang2025structured3dlatentsscalable}    & 0.0391 & 0.4771 & 0.4526 & 0.4605 & 0.8302 & 0.6486 \\
& SAM 3D  \cite{chen2025sam}   & 0.0296 & 0.5311 & \underline{0.7044} & 0.5872 & 0.8535 & 0.6339 \\
& ReconViaGen  \cite{chang2025reconviagen}  & \underline{0.0290} & \underline{0.6167} & 0.6149 & \underline{0.6069} & \underline{0.8751} & \underline{0.6918} \\
& \textbf{Ours}  & \textbf{0.0192} & \textbf{0.7322} & \textbf{0.7540} & \textbf{0.7384} & \textbf{0.9023} & \textbf{0.8155} \\
\bottomrule
\end{tabular}
\vspace{-2mm}
\end{table*}

\section{Experiments}
\subsection{Implementation Details}

\vspace{1mm}
\noindent\textbf{Data Preparation.} Our model is trained on a curated subset of Objaverse-XL~\cite{deitke2023objaverse}. Following the filtering pipeline in TRELLIS~\cite{xiang2025structured3dlatentsscalable}, we select models based on aesthetic scores and further exclude objects with transparent materials, resulting in a final training set of 40K high-quality 3D models. For each object, we render 50 viewpoints from randomly sampled camera poses using Blender to obtain multi-view RGB and depth images.
For evaluation, we select 100 objects from each of two widely-used benchmarks: Google Scanned Objects (GSO)~\cite{downs2022google}, which consists of high-quality scanned household items with diverse geometries and realistic textures, and Toys4k~\cite{stojanov2021using}, a collection of high-quality 3D toy meshes across different categories. For each object, we synthesize 24 multi-view RGB images across a diverse range of field-of-views, elevations, and azimuths. To simulate realistic conditions, we apply random perturbations to the camera translations. From this set of 24 views, we randomly select a 4-view subset as input to our model, allowing us to evaluate its performance in a challenging sparse-view reconstruction setting.

\vspace{1mm}
\noindent\textbf{Network Training.}
We train our framework in two stages for stability and efficiency. In Stage~I, we adapt the feed-forward multi-view reconstruction module to produce generation-compatible canonical predictions. We follow the standard training recipe of \cite{wang2025vggt} with two modifications: (1) we omit the tracking loss to simplify the objective and improve throughput, and (2) for each iteration, we randomly sample 4 views from the 50 rendered viewpoints per object. Training runs for 80K steps with a peak learning rate of $1\times 10^{-5}$ and a 1K-step linear warmup, using 4 NVIDIA H800 GPUs with a total batch size of 32 (about 5 days).
In Stage~II, we freeze the reconstruction module and train the 3D generator using the diffusion loss in Eq.~\eqref{eq:diffusion}. We sample ground-truth point clouds on-the-fly; to reduce supervision ambiguity caused by residual misalignment between ground truth and the predicted canonical geometry, we align the ground-truth point cloud to the predicted geometry with a similarity transform before encoding it into the latent space. The generator is trained for 80K steps with a peak learning rate of $1\times 10^{-5}$ and a 1K-step warmup on 4 NVIDIA H800 GPUs with a batch size of 32 over 8 days.
At inference, four input views take ${\sim}40$s per object at ${\sim}16$GB peak memory on one RTX A6000, versus ${\sim}27$s for the baseline Hunyuan3D-Omni; a detailed cost breakdown is given in supplementary Sec.~S11.

\vspace{1mm}
\noindent\textbf{Evaluation Metrics.}
We evaluate camera poses using \textit{Absolute Trajectory Error} (ATE) and \textit{Relative Pose Error} (RPE), reporting both translation ($RPE_{tr}$) and rotation ($RPE_{r}$) components. Depth accuracy is measured by \textit{Absolute Relative Error} (Abs~Rel) and \textit{Root Mean Square Error} (RMSE), and mesh geometry by \textit{Chamfer-$L_2$ Distance} (CD), \textit{Precision}, \textit{Recall}, and \textit{F-Score} at distance threshold $d<0.01$, \textit{Normal Consistency} (NC), and \textit{Voxel IoU} on a $128^3$ grid. For all methods, including ours, predictions are first aligned to the ground truth with an identical ICP procedure before computing the mesh metrics.

\subsection{Comparison with State-of-the-Art Methods}

\vspace{1mm}
\noindent\textbf{Multi-view Camera Pose and Depth Estimation.} 
We first evaluate our method on multi-view camera pose and depth estimation using the GSO and Toys4k datasets. We benchmark our model against the original VGGT baseline \cite{wang2025vggt} and ReconViaGen \cite{chang2025reconviagen}. We choose ReconViaGen for comparison as it represents the most closely related prior work to our approach; specifically, it is also built upon the VGGT architecture and leverages generative models to enhance 3D object reconstruction. The quantitative results are detailed in Table~\ref{tab:camera_pose} and Table~\ref{tab:depth}. While both our method and ReconViaGen significantly outperform the original VGGT after fine-tuning on 3D object datasets, our approach achieves superior performance. This demonstrates the effectiveness of our generation-compatible framework in achieving accurate geometry and camera poses.

\begin{table}[t]
    \centering
    \caption{\textbf{Quantitative Comparison of Camera Pose Estimation.} We evaluate Absolute Trajectory Error (ATE) and Relative Pose Error (RPE) for translation ($t$) and rotation ($r$) on the GSO and Toys4k benchmarks.}
    \vspace{-1em}

    \resizebox{1.0\columnwidth}!{
    \begin{tabular}{lcccccc}
        \toprule
        \multicolumn{1}{l}{\multirow{3}{*}{\textbf{Method}}} &
        \multicolumn{3}{c}{\textbf{GSO}} &
        \multicolumn{3}{c}{\textbf{Toys4k}} \\
        \cmidrule(r){2-4} \cmidrule(r){5-7}
        \multicolumn{1}{c}{} &
        ATE $\downarrow$ & RPE$_t$ $\downarrow$ & RPE$_r$ $\downarrow$ &
        ATE $\downarrow$ & RPE$_t$ $\downarrow$ & RPE$_r$ $\downarrow$ \\
        \midrule
        VGGT \cite{wang2025vggt} & 0.0799 & 0.1856 & 8.18 & 0.1209 & 0.2931 & 19.14 \\
        ReconViaGen \cite{chang2025reconviagen} & \underline{0.0190} & \underline{0.0443} & \textbf{2.12} & \underline{0.0425} & \underline{0.1099} & \underline{8.53} \\
        \textbf{Ours} & \textbf{0.0151} & \textbf{0.0338} & \underline{2.16} & \textbf{0.0255} & \textbf{0.0637} & \textbf{3.67} \\
        \bottomrule
    \end{tabular}}
    \vspace{-1.5em}
    \label{tab:camera_pose}
\end{table}

\vspace{1mm}
\noindent\textbf{Multi-view Consistent 3D Object Generation.}
We evaluate our framework against a diverse set of state-of-the-art 3D reconstruction and generative models. Specifically, we compare against TRELLIS-S \cite{xiang2025structured3dlatentsscalable}, which utilizes a stochastic denoising process by randomly selecting a single input view for conditioning at each step, and TRELLIS-M \cite{xiang2025structured3dlatentsscalable}, which employs a multidiffusion approach to aggregate and average the denoised results from all available input views. We further include Hunyuan3D-MV \cite{zhao2025hunyuan3d20scalingdiffusion}, a multi-view Diffusion Transformer (DiT) designed for geometric synthesis, and LucidFusion \cite{he2024lucidfusion}, a feed-forward model based on Relative Coordinate Maps where we employ the LGM  \cite{tang2024lgm} framework for mesh extraction. For single-view foundations, we compare with SAM 3D \cite{chen2025sam}; given its single-image constraint, we report the best performance among four independent per-view runs. Finally, we compare with ReconViaGen \cite{chang2025reconviagen}, a prior method that couples sparse-view reconstruction with generative priors. All baselines use their officially released weights without retraining; our model is trained on 40K Objaverse-XL objects, versus 390K for ReconViaGen.

\begin{table}[t]
    \centering
    \caption{\textbf{Quantitative Comparison of Depth Estimation.} Our method demonstrates superior accuracy in depth prediction compared to other baselines on both GSO and Toys4k datasets.}
    \vspace{-1em}
    \resizebox{1.0\columnwidth}!{
    \begin{tabular}{lcccc}
        \toprule
        \multicolumn{1}{l}{\multirow{3}{*}{\textbf{Method}}} &
        \multicolumn{2}{c}{\textbf{GSO}} &
        \multicolumn{2}{c}{\textbf{Toys4k}} \\
        \cmidrule(r){2-3} \cmidrule(r){4-5}
        \multicolumn{1}{c}{} &
        Abs Rel $\downarrow$ & RMSE $\downarrow$ &
        Abs Rel $\downarrow$ & RMSE $\downarrow$ \\
        \midrule
        VGGT \cite{wang2025vggt} & 0.0512 & 0.0686 & 0.0522 & 0.0735 \\
        ReconViaGen \cite{chang2025reconviagen} & \underline{0.0051} & \underline{0.0081} & \underline{0.0065} & \underline{0.0109} \\
        \textbf{Ours} & \textbf{0.0033} & \textbf{0.0057} & \textbf{0.0039} & \textbf{0.0075} \\
        \bottomrule
    \end{tabular}
    }
    \vspace{-1.5em}
    \label{tab:depth}
\end{table}

\begin{figure*}[p]
  \centering
  \vspace*{1.8cm} 

  \includegraphics[width=\textwidth]{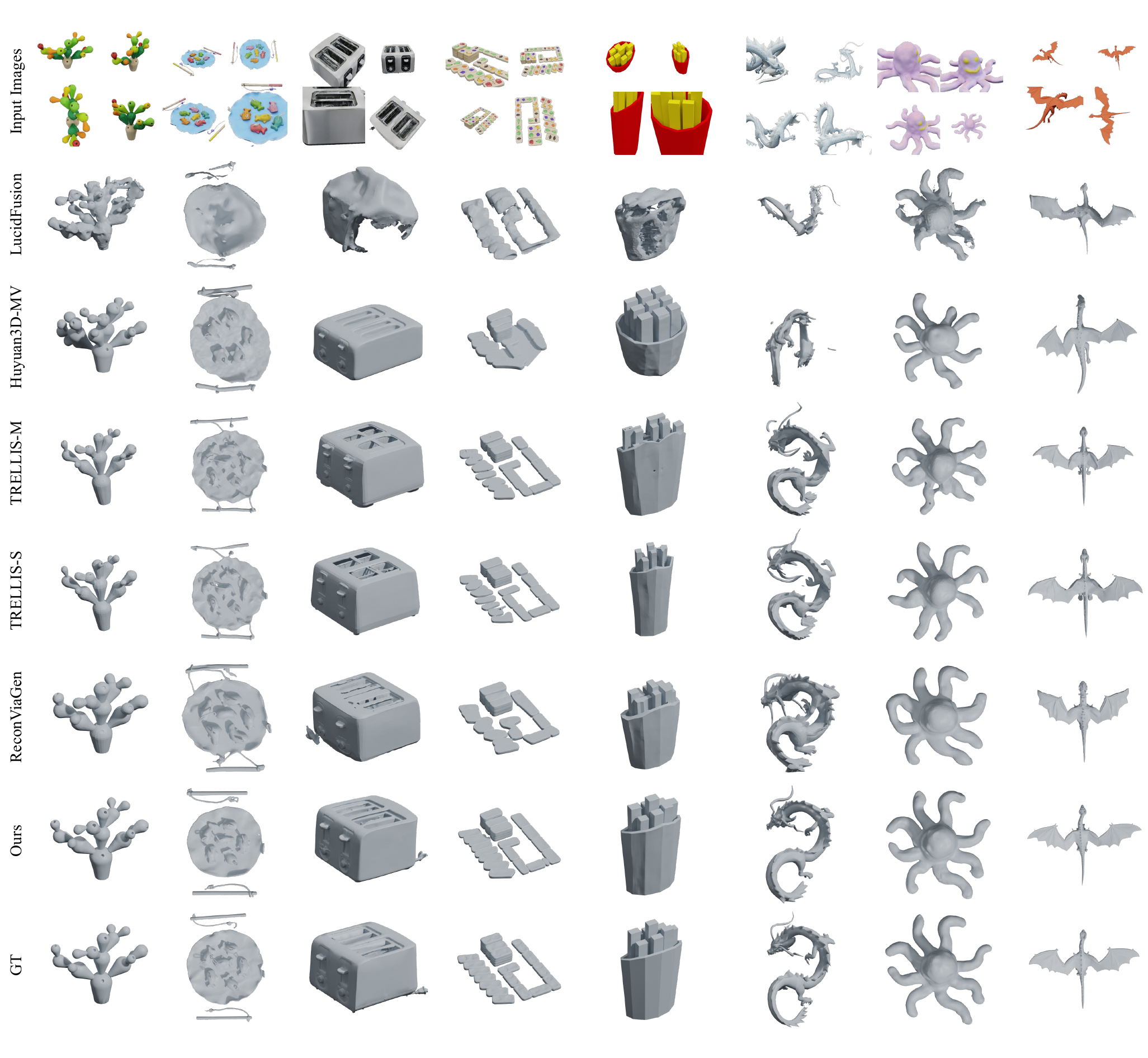}

  \caption{\textbf{Qualitative Comparison on Toys4K and GSO.} Compared to state-of-the-art reconstruction and generative baselines, our method produces 3D meshes with higher structural fidelity and superior multi-view consistency from sparse inputs.}
  \label{fig:qualitative_results}

  \vspace*{\fill} 
\end{figure*}

\begin{figure*}[t]
\vspace{-0.2cm}
  \centering
  \includegraphics[width=1.0\linewidth]{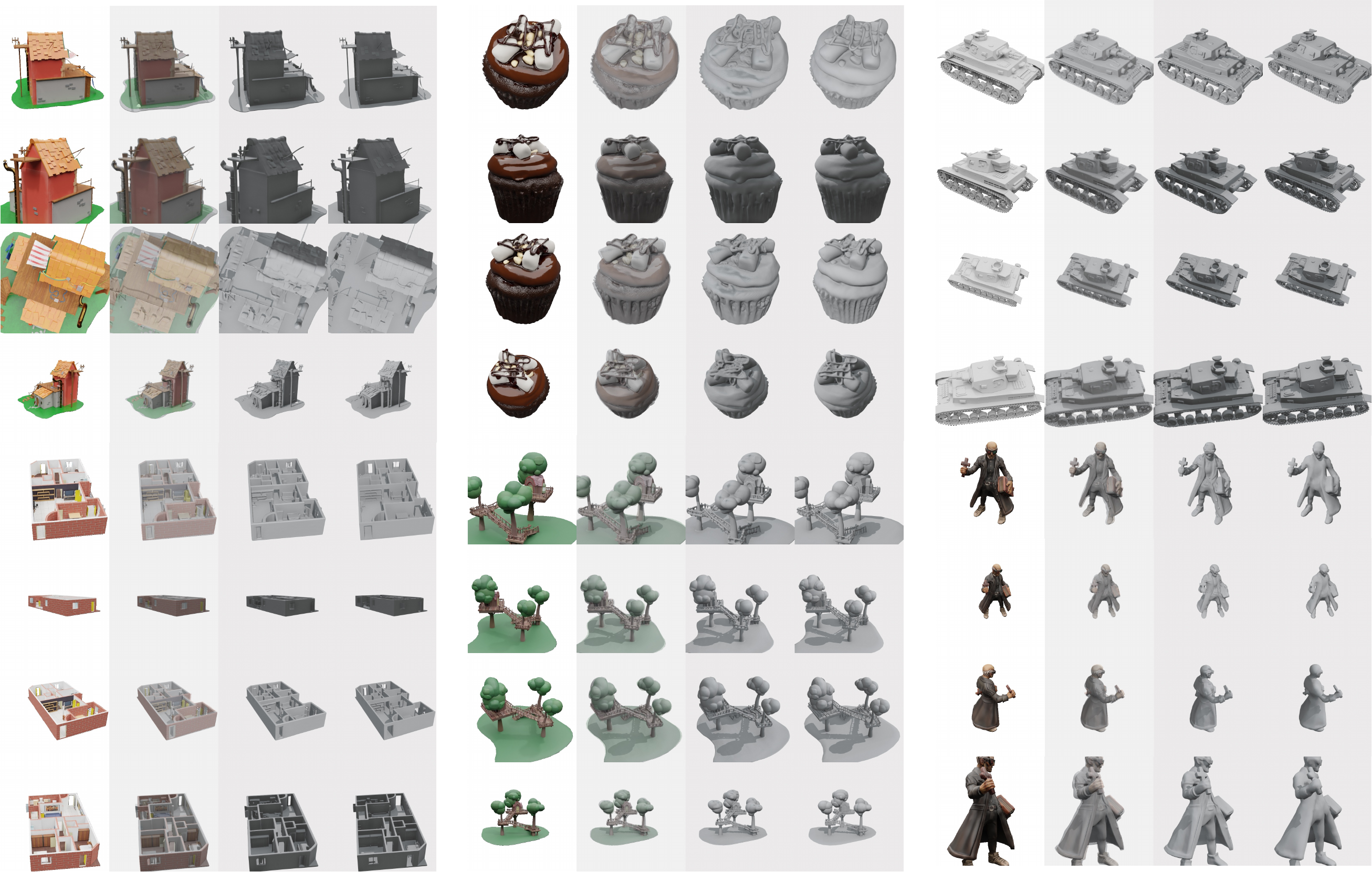}
  \vspace{-0.8cm}
\caption{\textbf{Additional qualitative results.} We show six multi-view shape modeling examples. For each of four input views, from left to right we show: the input image, our mesh rendering overlaid on the input image, our mesh rendering, and the ground-truth mesh rendering.}
\label{fig:more}

    \vspace{-0.3cm}
\end{figure*}

As shown in Table~\ref{tab:main_results} and Fig.~\ref{fig:qualitative_results}, our method consistently outperforms all baselines across both Toys4K and GSO datasets. Notably, LucidFusion exhibits a significant performance drop under our benchmark. This stems from its reliance on per-view geometry estimation, which is inherently unable to reconstruct unseen regions and struggles with unposed or perturbed camera translations in our evaluation.
Furthermore, our framework demonstrates superior performance across all geometric metrics when compared to ReconViaGen. While ReconViaGen relies on injecting implicit features from a reconstruction branch into the generative process, our method provides explicit guidance by informing the generation model with intermediate reconstructed geometry. This strategy allows the model to better resolve spatial ambiguities and maintain stricter structural fidelity, suggesting that direct geometric conditioning is a more effective prior than high-level feature injection; two controlled studies in the supplementary material further isolate the source of this gain (Secs.~S6 and~S7).
These quantitative gains are further supported by the qualitative results in Fig.~\ref{fig:qualitative_results}, demonstrating that our method outperforms other state-of-the-art baselines in both mesh quality and multi-view consistency. Our approach consistently delivers high-fidelity 3D object modeling results.
As illustrated in Fig.~\ref{fig:real_world}, our model generalizes effectively to real-world sparse-view inputs, consistently achieving superior multi-view consistency and mesh quality.
Quantitatively, our method also outperforms ReconViaGen on real casual captures with cluttered backgrounds on the CO3D benchmark (supplementary Sec.~S5).
More qualitative results can be found in Fig.~\ref{fig:more}.

\begin{figure}[t]
    \centering
    \includegraphics[width=\linewidth]{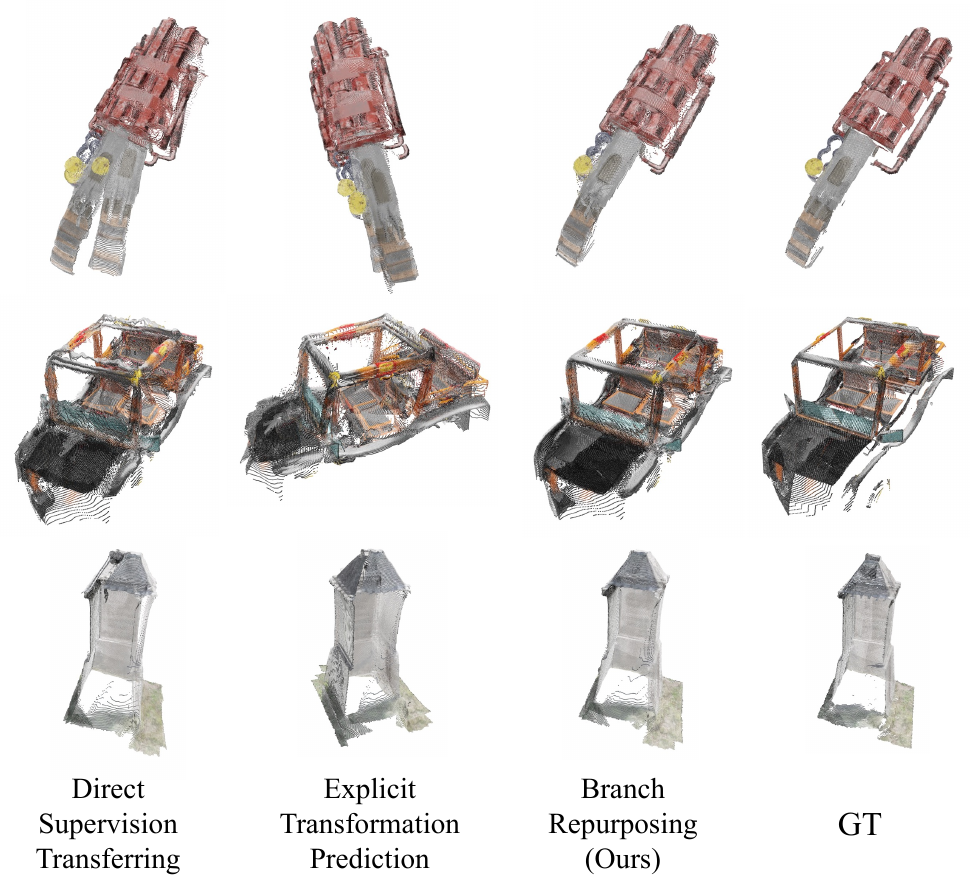}
    \vspace{-8mm}
    \caption{\textbf{Qualitative Comparison of Canonical Alignment.} Our strategy achieves better geometric quality while aligning the canonical object space.}
    \label{fig:ablation_results_cam}
    \vspace{-4mm}
\end{figure}
\begin{table}[t]
    \centering
    \caption{\textbf{Quantitative Comparison of Canonical Alignment.} Compared with other strategies, our  strategy is more accurate in camera pose prediction while remaining comparable in depth estimation.}
    \label{tab:ablation}
    \vspace{-1em}
    \resizebox{1.0\columnwidth}!{
    \begin{tabular}{lccccc}
        \toprule
        \multirow{2}{*}{\textbf{Strategy}} & \multicolumn{3}{c}{\textbf{Camera Pose}} & \multicolumn{2}{c}{\textbf{Depth}} \\
        \cmidrule(r){2-4} \cmidrule(r){5-6}
        & ATE $\downarrow$ & RPE$_t$ $\downarrow$ & RPE$_r$ $\downarrow$ & Abs Rel $\downarrow$ & RMSE $\downarrow$ \\
        \midrule
        Direct Supervision Transferring   & \underline{0.2517} & \underline{0.1414} & \underline{4.2376} & 0.0062 & 0.0210 \\
        Explicit Rigid Transformation & 0.5934 & 0.1474 & 4.4596 & \textbf{0.0049} & \textbf{0.0191} \\
        \textbf{Branch Repurposing (Ours)} & \textbf{0.0353} & \textbf{0.0453} & \textbf{1.2181} & \underline{0.0054} & \underline{0.0200} \\
        \bottomrule
    \end{tabular}}
    \vspace{-1.5em}
\end{table}

\subsection{Ablation Studies}
Due to the significant computational requirements of the full training cycle, we conduct our ablation studies on a reduced experimental scale. Specifically, we train the models using a subset of 2,000 objects sampled from our primary training set and evaluate performance on a held-out test set consisting of 50 diverse objects.
The supplementary material adds a joint-training comparison quantifying the disparate-dynamics challenge (Sec.~S8) and a $\pi^3$ backbone swap (Sec.~S9).

\vspace{1mm}
\noindent\textbf{Canonical Alignment Strategy.}
A central challenge in our framework is reconciling the camera-centric reconstruction model with the canonical object space required by the diffusion generator. To address this, we compare our selective supervision strategy against two alternative approaches: explicit rigid transformation regression and direct supervision transferring. 
Quantitative and qualitative comparisons (Table~\ref{tab:ablation} and Fig.~\ref{fig:ablation_results_cam}) highlight the trade-offs of each method. 
While explicit regression achieves marginally better depth metrics, it suffers from the poorest global $SE(3)$ alignment, as evidenced by a significantly higher ATE. This supports our hypothesis that regressing a global pose provides a weaker supervision signal for spatial alignment than dense, pointmap-based supervision. Conversely, direct supervision transferring leads to a marked performance decay in both camera pose and depth estimation. This suggests that enforcing a global coordinate shift without an adaptation mechanism disrupts the model's pre-trained 3D priors. In contrast, our branch repurposing strategy achieves an optimal balance: it facilitates precise camera alignment and high-fidelity depth estimation by effectively canonicalizing the reference frame while preserving essential geometric priors.

\begin{figure}[t]
    \centering
    \includegraphics[width=\linewidth]{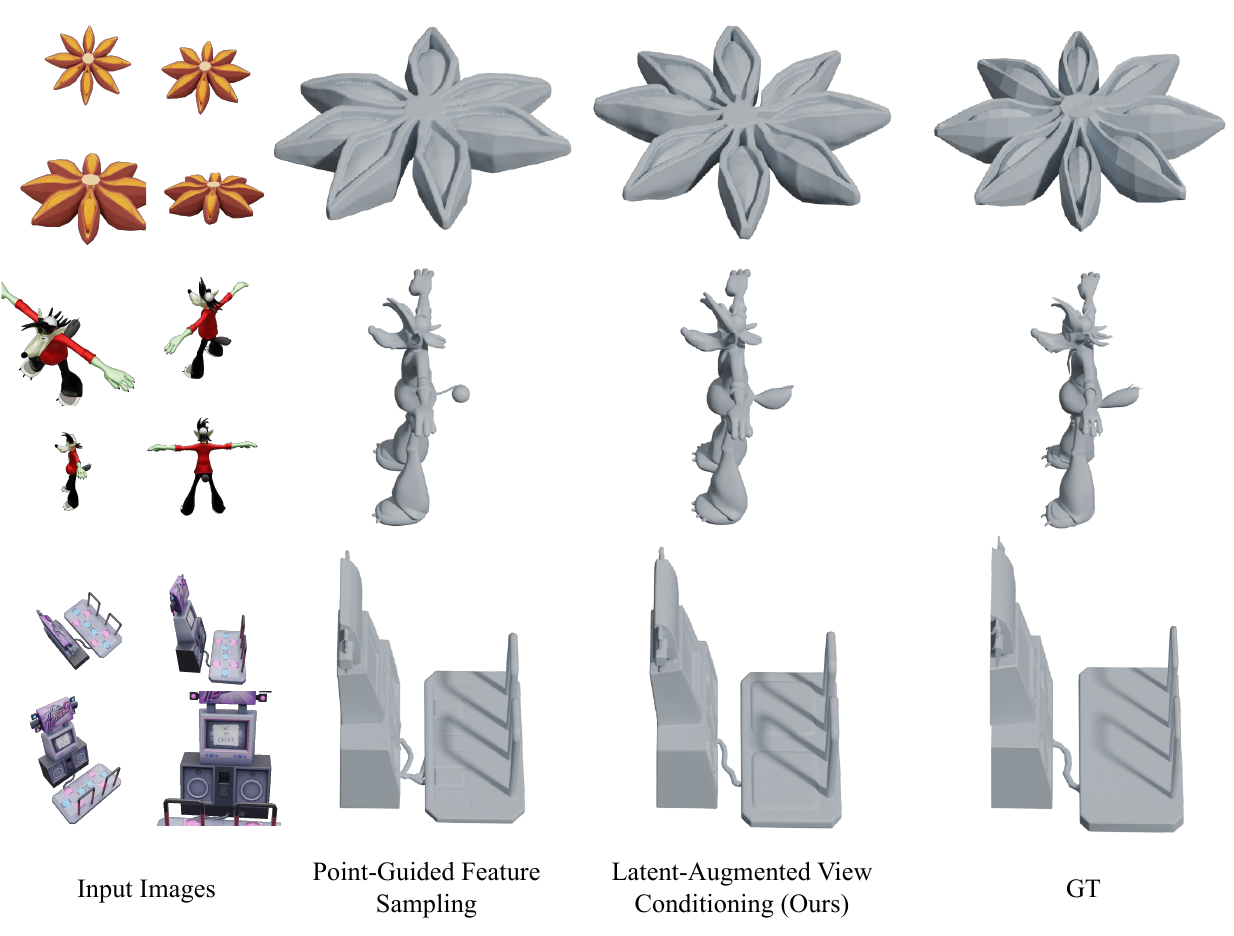}
    \vspace{-8mm}
    \caption{\textbf{Qualitative Comparison of Multi-view Condition.} Our strategy (right) preserves dense image context more effectively than point-guided sampling (left), leading to improved input alignment.}
    \label{fig:ablation_results_3d}
    \vspace{-4mm}
\end{figure}
\begin{table}[t]
\centering
\caption{\textbf{Quantitative Comparison of Multi-view Condition.} Our strategy achieves the best performance across most geometric metrics.}
\label{tab:ablation_conditioning}
\vspace{-1em}
\resizebox{\columnwidth}{!}{
\begin{tabular}{lcccccc}
\toprule
Strategy & Chamfer-$L_2 \downarrow$ & Prec. $\uparrow$ & Rec. $\uparrow$ & F-Score $\uparrow$ & Normal $\uparrow$ & IoU $\uparrow$ \\
\midrule
Point-Guided Sampling & 0.0260 & 0.7677 & 0.7087 & 0.7208 & \textbf{0.8419} & 0.6304 \\
Latent-Augmented (Ours) & \textbf{0.0259} & \textbf{0.7890} & \textbf{0.7325} & \textbf{0.7428} & 0.8350 & \textbf{0.6707} \\
\bottomrule
\end{tabular}}
\vspace{-1.5em}
\end{table}

\vspace{1mm}
\noindent\textbf{Multi-view Conditioning Strategy.}
We evaluate our multi-view integration strategy by comparing our latent-augmented view conditioning against the point-guided sampling baseline introduced in Sec.~\ref{sec:method_gen}.
Quantitative results in Table~\ref{tab:ablation_conditioning} demonstrate that our approach leads to better 3D shape modeling performance.
Qualitatively, Fig.~\ref{fig:ablation_results_3d} shows that the point-guided baseline can result in minor artifacts and inconsistencies.
We attribute this to the fact that indexing image features with sparse points causes a loss of the original dense image context, which limits the model's ability to maintain multi-view consistency. In contrast, our method preserves the complete dense feature set while augmenting it with geometric context.
This conclusion is stable across three inference runs with different random seeds (supplementary Sec.~S10).

\section{Limitations and Future Work}
\label{sec:limitations}

\begin{figure}[t]
    \centering
        \includegraphics[width=\linewidth]{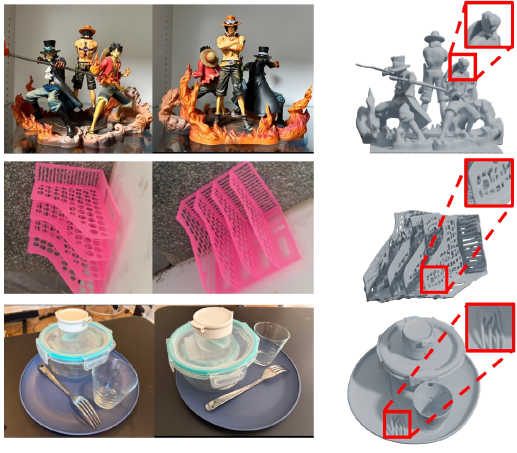}
    \vspace{-6mm}
    \caption{\textbf{Failure cases of our full method.} Top to bottom: mutual occlusion loses fine facial detail; thin, perforated structures are partially recovered; reflective objects are less robust under real-world illumination.}
    \label{fig:limitations}
    \vspace{-4mm}
\end{figure}

\begin{figure*}[t]
  \centering
  \vspace{-0.6cm}
  \includegraphics[width=1.0\linewidth]{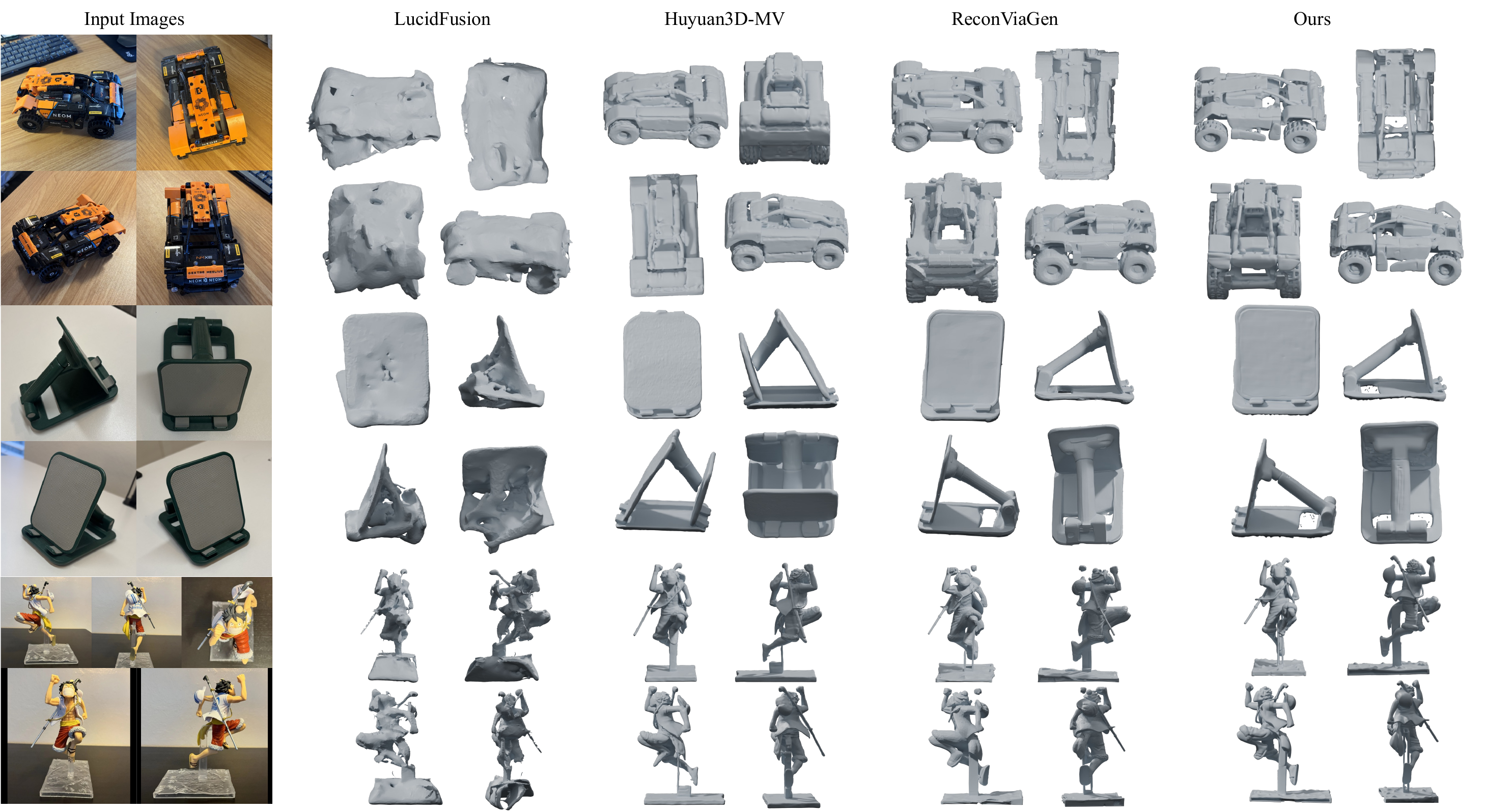}
  \vspace{-0.7cm}
  \caption{\textbf{Generalization to Real-world Environments.} Our framework demonstrates robustness and superior performance compared to SOTA method.}
  
  \label{fig:real_world}
\end{figure*}

While RecGen3D recovers faithful geometry from sparse unposed captures, its robustness is ultimately bounded by the synthetic training data. The method generalizes well to casual real photographs (Fig.~\ref{fig:real_world}, supplementary Sec.~S5), yet heavy occlusion and reflective materials under uncontrolled illumination still lead to degraded reconstructions (Fig.~\ref{fig:limitations}). In addition, the generative prior appears to favor smooth, well-connected surfaces: very thin or densely perforated structures may be simplified or broken, reducing fidelity to fine input details. We expect training on real multi-view captures and stronger detail-preserving conditioning to mitigate both issues, and plan to extend the framework toward scene-level modeling and texture synthesis.

\section{Conclusion}
\noindent
We present RecGen3D, a reconstruct-then-condition framework that combines feed-forward multi-view reconstruction with controllable 3D diffusion to recover high-fidelity object geometry from unposed images.
Our approach bridges the coordinate and representation gap between the two paradigms by canonicalizing reconstruction outputs and using them as explicit geometric guidance for multi-view conditioned generation.
Extensive experiments show consistent improvements over strong reconstruction and generative baselines on Toys4K and GSO, and demonstrate robust generalization to real-world captures.
By harmonizing these distinct methodologies, we aim to provide a foundation for the close integration of 3D reconstruction and generative modeling, fostering more integrated and versatile 3D vision systems.

\bibliographystyle{ACM-Reference-Format}
\bibliography{hzs}

\clearpage
\raggedbottom
\setcounter{section}{0}
\setcounter{figure}{0}
\setcounter{table}{0}
\setcounter{equation}{0}
\renewcommand{\thesection}{S\arabic{section}}
\renewcommand{\thefigure}{S\arabic{figure}}
\renewcommand{\thetable}{S\arabic{table}}
\renewcommand{\theequation}{S\arabic{equation}}
\begin{center}
  {\LARGE\bfseries Supplementary Material}
\end{center}
\vspace{1em}

\section*{Overview}
\label{sec:supp_overview}

This supplementary material provides additional details and results that
complement the main paper. The content is organized as follows:

\begin{itemize}
    \item \textbf{Sec.~\ref{sec:supp_ablation_components}} reports a
    component-wise ablation that isolates the contribution of branch
    repurposing and VGGT feature injection
    (Table~\ref{tab:ablation_components}).

    \item \textbf{Sec.~\ref{sec:supp_qual_vggt}} presents qualitative
    comparisons that visualize the failure modes of removing VGGT feature
    injection and the structural advantages of our full design
    (Fig.~\ref{fig:ablation_vggt}).

    \item \textbf{Sec.~\ref{sec:supp_generalization}} evaluates the
    generalization of our framework to varying numbers of input views
    (2~views and 8~views), with quantitative results in
    Table~\ref{tab:generalization_views} and extensive qualitative
    comparisons against ReconViaGen in Fig.~\ref{fig:generalization_2view}
    and Fig.~\ref{fig:generalization_8view}.

    \item \textbf{Sec.~\ref{sec:supp_in_the_wild}} demonstrates the
    robustness of \method on in-the-wild captures with $2$ to $16$
    unposed views collected under uncontrolled real-world conditions
    (Fig.~\ref{fig:in_the_wild}).

    \item \textbf{Sec.~\ref{sec:supp_co3d}} quantitatively evaluates
    real-world generalization on the CO3D benchmark
    (Table~\ref{tab:co3d}, Fig.~\ref{fig:co3d_qual}).

    \item \textbf{Sec.~\ref{sec:supp_ladder}} presents a
    source-of-gain analysis that conditions the untouched pretrained
    generator on progressively stronger reconstruction inputs
    (Table~\ref{tab:ladder}).

    \item \textbf{Sec.~\ref{sec:supp_backbone_matched}} compares our
    conditioning design against ReconViaGen's implicit feature injection
    under a matched generative backbone, data, and training budget
    (Table~\ref{tab:backbone_matched}).

    \item \textbf{Sec.~\ref{sec:supp_joint}} quantifies the
    disparate-learning-dynamics challenge with a controlled joint-training
    comparison (Table~\ref{tab:joint_training}).

    \item \textbf{Sec.~\ref{sec:supp_pi3}} verifies that our recipe
    transfers to a different reconstruction backbone, $\pi^3$
    (Table~\ref{tab:pi3_swap}, Fig.~\ref{fig:pi3_qual}).

    \item \textbf{Sec.~\ref{sec:supp_multirun}} reports a multi-run
    version of the conditioning ablation with mean and standard deviation
    over three inference runs with different random seeds
    (Table~\ref{tab:multirun}).

    \item \textbf{Sec.~\ref{sec:supp_impl}} provides additional
    implementation details: conditioning-encoder architectures, sampling
    settings, the multi-view token layout, and an inference-cost
    breakdown.
\end{itemize}

We strongly encourage readers to view the accompanying supplementary
figures at full resolution, as fine-grained geometric differences are
most evident in zoomed-in surface details.

\section{Component-wise Ablation Study}
\label{sec:supp_ablation_components}

To complement the main-paper ablations on the canonical alignment
strategy and the multi-view conditioning strategy, we present an
additional component-wise study that simultaneously isolates the
contributions of \emph{branch repurposing} and \emph{VGGT feature
injection}, the two central design decisions of our framework.
Table~\ref{tab:ablation_components} reports the results under three
configurations: (i)~removing branch repurposing (replacing it with
direct supervision transferring, while keeping VGGT feature injection),
(ii)~removing VGGT feature injection (while keeping branch
repurposing), and (iii)~our full method that integrates both
components. All variants are trained under the same reduced
experimental protocol described in the main paper for ablation studies.

\begin{table}[t]
\centering
\caption{\textbf{Component-wise ablation study.} We isolate the
contribution of \emph{Branch Repurposing} (canonical alignment) and
\emph{VGGT Feature Injection} (multi-view geometric conditioning).
\cmark{} indicates the component is enabled. Removing either
component leads to consistent degradation across all geometric
metrics, confirming that both designs are necessary and complementary.}
\label{tab:ablation_components}
\vspace{-1em}
\resizebox{\linewidth}{!}{
\begin{tabular}{lcccccc}
\toprule
\textbf{Method}
& \textbf{Branch}
& \textbf{VGGT}
& \textbf{CD$\downarrow$}
& \textbf{Prec.$\uparrow$}
& \textbf{Rec.$\uparrow$}
& \textbf{F-Score$\uparrow$} \\
& \textbf{Repurposing}
& \textbf{Features}
& & & & \\
\midrule
w/o Branch Repurposing
& \xmark & \cmark
& 0.0339 & 0.7229 & 0.6407 & 0.6535 \\
w/o VGGT Features
& \cmark & \xmark
& 0.0301 & 0.7189 & 0.6711 & 0.6761 \\
\textbf{Full Method (Ours)}
& \cmark & \cmark
& \textbf{0.0259} & \textbf{0.7890} & \textbf{0.7325} & \textbf{0.7428} \\
\bottomrule
\end{tabular}
}
\vspace{-0.5em}
\end{table}

Removing branch repurposing causes the largest drop in
Chamfer-$L_2$ and F-Score, confirming that accurate canonical
alignment is critical: when the reconstruction branch is supervised
directly in canonical space without branch repurposing, the pretrained
3D priors of VGGT are disrupted, and the geometric anchor passed to
the generator becomes unreliable. Removing VGGT feature injection also
degrades all metrics, demonstrating that geometric tokens provide
essential spatial grounding beyond what multi-view appearance features
alone can offer. The two designs are complementary: branch repurposing
guarantees that the geometric anchor lives in the canonical space the
generator expects, while VGGT feature injection equips the diffusion
generator with continuous geometric context that disambiguates
multi-view appearance.

\section{Qualitative Ablation: VGGT Feature Injection}
\label{sec:supp_qual_vggt}

\begin{figure*}[t]
  \centering
  \includegraphics[width=\linewidth]{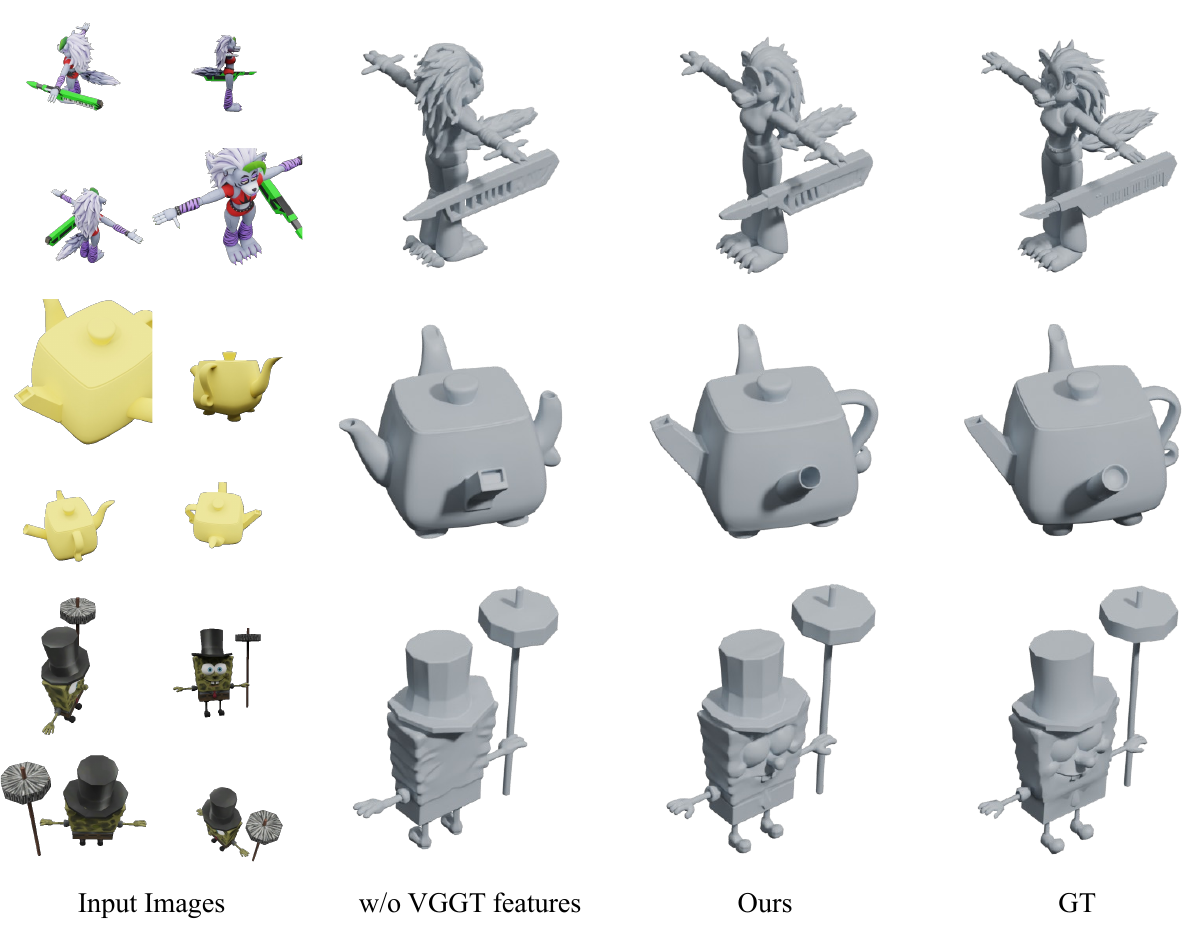}
  \caption{\textbf{Qualitative comparison of VGGT feature injection.}
  From left to right: input images, results without VGGT features,
  our full method, and ground truth. Without VGGT geometric tokens,
  the generator relies solely on DINO appearance features and fails
  to disambiguate geometrically distinct viewpoints. This manifests
  as front-back inversion in articulated objects (rows~1 and 3) and
  geometric aliasing across structurally asymmetric surfaces (row~2,
  teapot). Our full method injects VGGT latent tokens as spatial
  anchors, resolving viewpoint ambiguity and preserving
  instance-specific structural fidelity.}
  \label{fig:ablation_vggt}
\end{figure*}

Figure~\ref{fig:ablation_vggt} provides qualitative comparisons
between the variant without VGGT feature injection and our full
method. Both variants share the same branch-repurposing strategy for
canonical alignment, but differ in whether VGGT latent tokens and
camera tokens are injected as geometric embeddings into the multi-view
conditioning signal.

The key failure mode of the w/o VGGT variant reveals the precise role
that geometric tokens play in our framework. As discussed in the main
paper, VGGT tokens $\mathbf{F}_i^{V}$ represent a continuous, learned
geometric manifold with global structural coherence, which enables the
diffusion model to precisely localize and distinguish semantic
information from different viewpoints within the established 3D
structure. Without this geometric grounding, the generator must rely
solely on multi-view DINO tokens, which capture rich appearance
semantics but lack explicit 3D spatial anchoring. This causes the
model to conflate features from geometrically distinct viewpoints. In
the first example (Fig.~\ref{fig:ablation_vggt}, row~1), the front
and back of the character are inverted---facial features appear on the
rear surface, while finger geometry is duplicated on both the front
and back faces. A similar view-confusion artifact is visible in the
third example (row~3), where structural elements from opposing views
are incorrectly merged. The effect is most pronounced in the teapot
example (row~2): without a geometric anchor to disambiguate the four
input viewpoints, the geometrically distinct spout, handle, lid, and
body surfaces collapse into a blended, aliased shape that fails to
recover the correct topology. Our full method, by injecting VGGT
tokens as structural anchors, provides the diffusion generator with a
continuous geometric scaffold that resolves viewpoint ambiguity,
preserves instance-specific structural features, and ensures
multi-view consistency throughout generation.

\section{Generalization to Varying Numbers of Input Views}
\label{sec:supp_generalization}

Although our framework is trained with a fixed number of $4$ input
views (as described in the main paper), real-world applications often
involve a variable number of unposed observations. To probe the
robustness of \method beyond its training regime, we evaluate it
under both \emph{sparser} ($2$-view) and \emph{denser} ($8$-view)
input settings on the GSO benchmark, and compare against ReconViaGen,
which represents the most closely related prior integration approach.

\paragraph{Quantitative results.}
Table~\ref{tab:generalization_views} reports the geometric metrics
under the two settings. \method consistently outperforms ReconViaGen
across all metrics in both regimes, with the relative gain being
particularly pronounced in the challenging $2$-view setting. This
indicates that the canonical geometric anchor produced by our
branch-repurposing module remains reliable even when the visual
evidence is highly incomplete, and that the latent-augmented
multi-view conditioning generalizes gracefully beyond the
training-time view count.

\begin{table}[t]
\centering
\caption{\textbf{Generalization to varying numbers of input views on
GSO.} Both methods are trained with $4$ input views. Our framework
maintains a consistent advantage over ReconViaGen across both the
sparse ($2$-view) and denser ($8$-view) regimes.}
\label{tab:generalization_views}
\vspace{-1em}
\resizebox{\linewidth}{!}{
\begin{tabular}{lcccccc}
\toprule
\textbf{Method}
& \textbf{CD$\downarrow$}
& \textbf{Prec.$\uparrow$}
& \textbf{Rec.$\uparrow$}
& \textbf{F-Score$\uparrow$}
& \textbf{Normal$\uparrow$}
& \textbf{IoU$\uparrow$} \\
\midrule
\multicolumn{7}{l}{\textit{$2$ input views}} \\
\midrule
ReconViaGen
& 0.0297 & 0.5311 & 0.5247 & 0.5222 & 0.8733 & 0.6992 \\
\textbf{Ours}
& \textbf{0.0221} & \textbf{0.6491} & \textbf{0.6850}
& \textbf{0.6622} & \textbf{0.8824} & \textbf{0.7805} \\
\midrule
\multicolumn{7}{l}{\textit{$8$ input views}} \\
\midrule
ReconViaGen
& 0.0234 & 0.6654 & 0.6567 & 0.6544 & 0.8881 & 0.7462 \\
\textbf{Ours}
& \textbf{0.0177} & \textbf{0.7569} & \textbf{0.7964}
& \textbf{0.7711} & \textbf{0.9064} & \textbf{0.8263} \\
\bottomrule
\end{tabular}
}
\vspace{-0.5em}
\end{table}

\begin{figure*}[t]
  \centering
  \includegraphics[width=\linewidth]{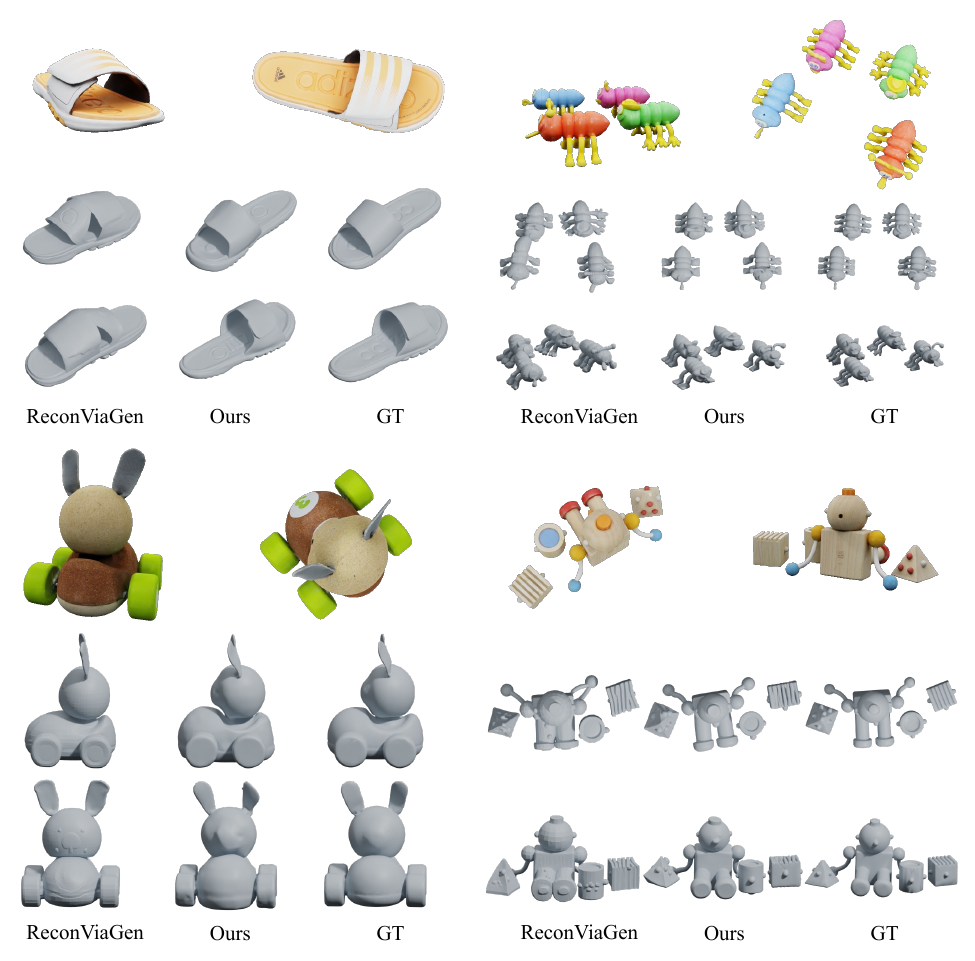}
  \caption{\textbf{Qualitative comparison with ReconViaGen under
  $2$ input views on GSO.} Each group shows ReconViaGen, ours, and
  ground truth rendered from two viewpoints. Under this challenging
  sparse-view setting, our method exhibits stronger generative
  completion: global shape structure is better preserved (row~1, gear
  puzzle; row~2, fingerless gloves) and fine geometric details remain
  coherent even when large portions of the object are unobserved
  (row~3, character with umbrella; row~4, airplane set). ReconViaGen
  tends to produce over-smoothed or topologically incorrect surfaces
  under such limited input.}
  \label{fig:generalization_2view}
\end{figure*}

\paragraph{Qualitative results under $2$ input views.}
Figure~\ref{fig:generalization_2view} compares our reconstructions
against ReconViaGen under the most challenging $2$-view setting,
where over half of the object surface is unobserved. In this regime,
the generative completion ability is critical: ReconViaGen tends to
produce over-smoothed or topologically incorrect surfaces, including
incorrect part counts in the gear puzzle (row~1), merged finger
geometry in the fingerless gloves (row~2), missing supports in the
character with umbrella (row~3), and disconnected parts in the
airplane set (row~4). In contrast, \method consistently preserves
global shape structure, recovers thin and articulated parts, and
yields fine geometric details even when large portions of the object
are invisible, thanks to the joint guidance of the canonical
geometric anchor and the generative prior.

\begin{figure*}[t]
  \centering
  \includegraphics[width=\linewidth]{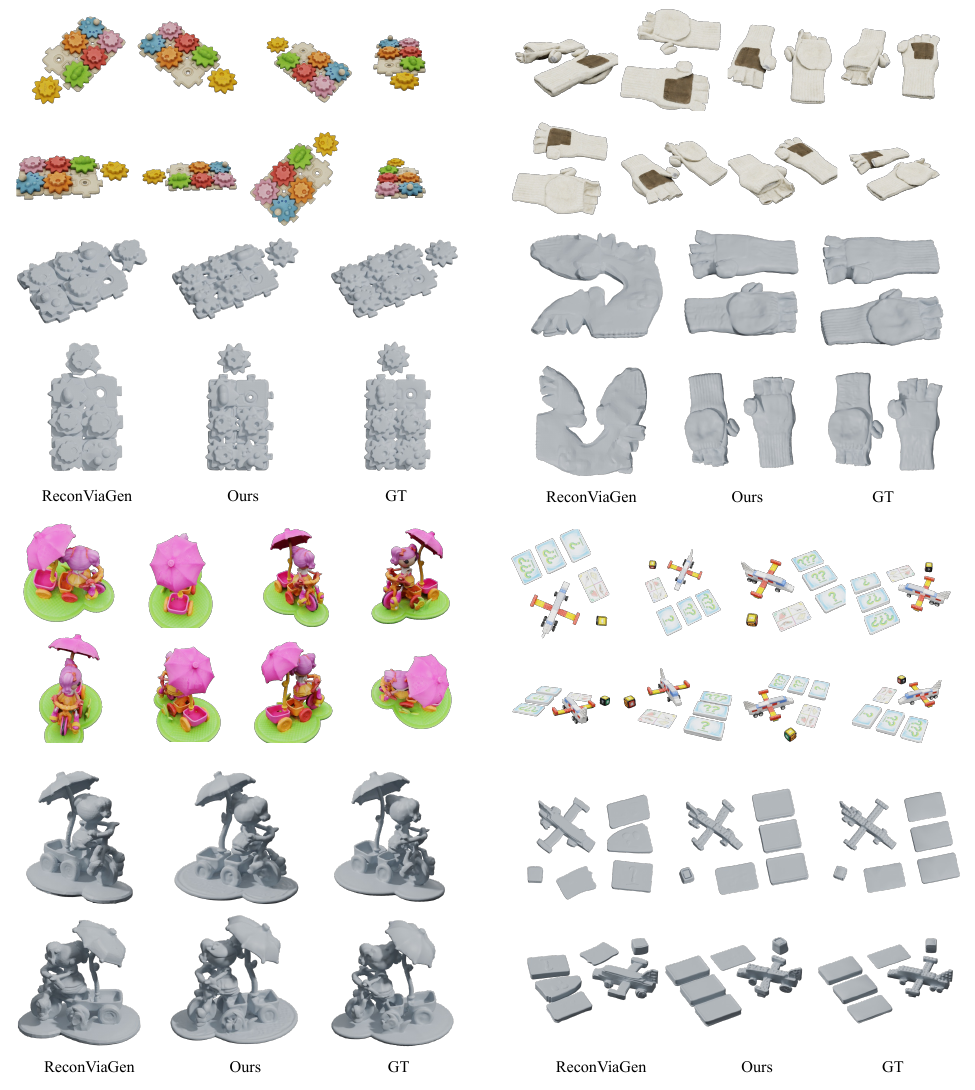}
  \caption{\textbf{Qualitative comparison with ReconViaGen under
  $8$ input views on GSO.} Each group shows ReconViaGen, ours, and
  ground truth rendered from two viewpoints. With denser input
  coverage, our method consistently produces more complete and
  geometrically faithful meshes, recovering fine-grained structures
  such as thin appendages (row~1, insect figurines), articulated
  components (row~2, bunny car), and complex part assemblies (row~4,
  robot toy) that ReconViaGen fails to reconstruct accurately.}
  \label{fig:generalization_8view}
\end{figure*}

\paragraph{Qualitative results under $8$ input views.}
Figure~\ref{fig:generalization_8view} shows reconstructions under
denser input coverage. With more views, \method exploits the
additional multi-view evidence to recover fine-grained structures
that remain difficult for ReconViaGen, including thin appendages
(row~1, insect figurines), articulated components (row~2, bunny car),
and complex part assemblies (row~4, robot toy). These observations
confirm that our framework scales gracefully with the amount of
visual evidence: it remains faithful to the observed surfaces while
exploiting denser views to refine geometric details.

\section{Additional In-the-Wild Qualitative Results}
\label{sec:supp_in_the_wild}

\begin{figure*}[h]
  \centering
  \includegraphics[width=\linewidth]{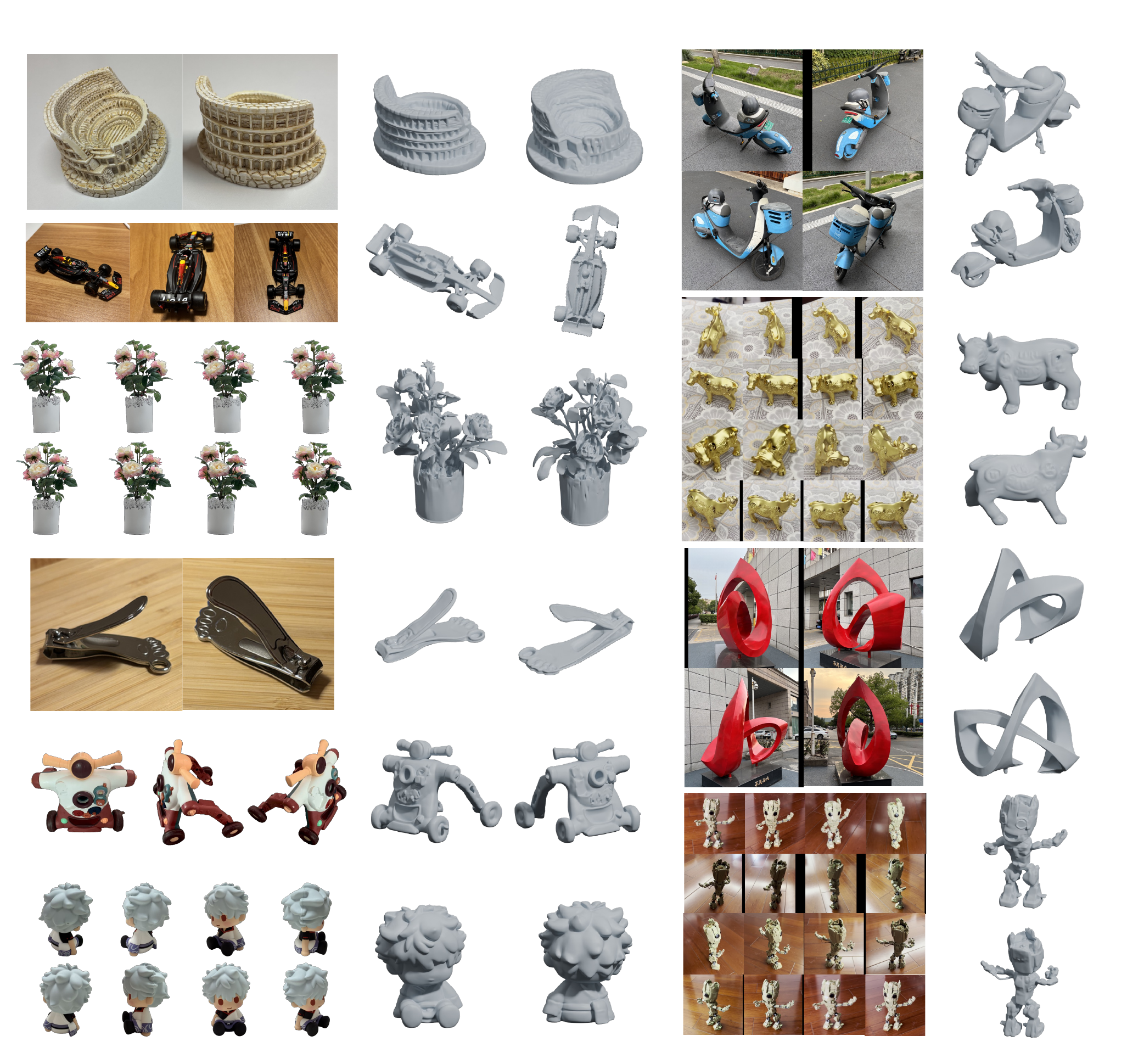}
  \caption{\textbf{Additional in-the-wild qualitative results.}
  Each group shows input images (left) and our reconstructed mesh
  (right). Input view counts range from $2$ to $16$ unposed images
  captured under uncontrolled real-world conditions. Despite being
  trained on fixed $4$-view synthetic data, \method generalizes
  robustly to varying view counts, uncontrolled camera poses, and
  diverse object categories, including architectural models,
  vehicles, organic objects, and articulated figurines.}
  \label{fig:in_the_wild}
\end{figure*}

To further demonstrate the practical applicability of \method beyond
controlled synthetic benchmarks, Fig.~\ref{fig:in_the_wild} presents
additional qualitative results on in-the-wild captures. Unlike the
evaluation setting in the main paper, these examples are sourced from
real-world photographs with \emph{no} ground-truth geometry,
uncontrolled lighting, cluttered backgrounds, and significant
appearance variation across views. Critically, the number of input
images is \emph{not} fixed: examples range from as few as $2$ to as
many as $16$ unposed views, and none of them shares a consistent
field of view or camera distance.

This setting stress-tests two core properties of our framework.
\emph{First}, the branch-repurposing strategy must reliably
canonicalize VGGT predictions regardless of the number of input
views and the degree of camera perturbation present in casually
captured images. \emph{Second}, the latent-augmented multi-view
conditioning must remain stable when the number of conditioning views
varies at inference time, a regime not explicitly seen during
training, where the model is exposed only to fixed $4$-view inputs.

The results show that \method generalizes gracefully to both sparse
and dense unposed captures. With as few as $2$ input images
(\emph{e.g.}, the nail clipper and the F1 car), the model recovers
plausible geometry for unobserved surfaces by leveraging the learned
generative prior, while remaining faithful to the visible evidence.
With denser inputs (\emph{e.g.}, the flower bouquet, the Colosseum
miniature, and the Groot figurine, captured from $12$--$16$
viewpoints), the framework exploits the richer multi-view coverage to
produce more complete and geometrically consistent reconstructions.
Across all examples, the canonical point cloud established by the
reconstruction module provides a reliable geometric scaffold that
guides the diffusion generator to produce high-fidelity meshes with
correct global structure, even under the challenging conditions of
in-the-wild capture.

\section{\texorpdfstring{Quantitative Real-World Evaluation on CO3D}{Quantitative Real-World Evaluation on CO3D}}
\label{sec:supp_co3d}

\begin{figure}[H]
  \centering
    \includegraphics[width=\linewidth]{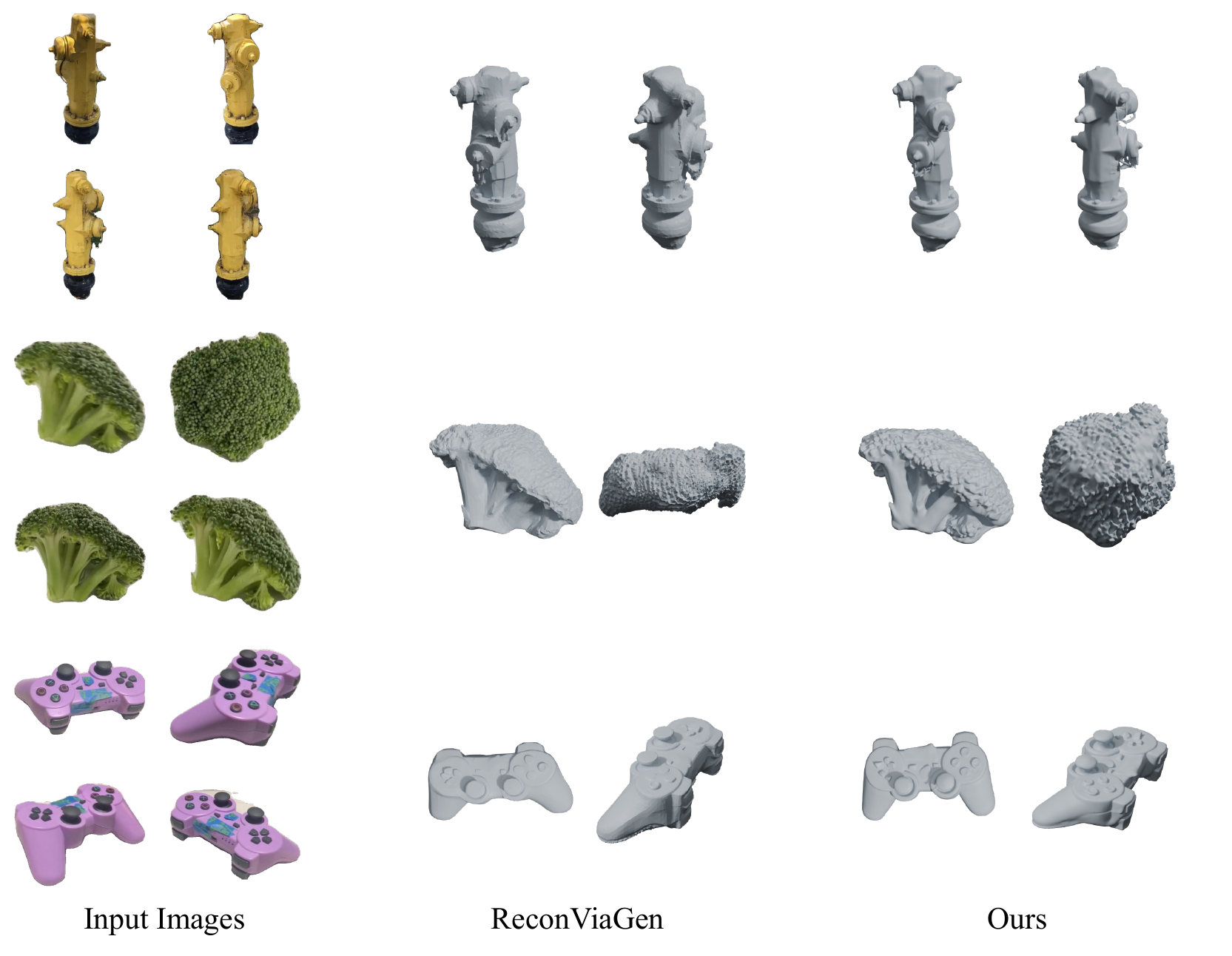}
  \caption{\textbf{Qualitative comparison on CO3D.} From four
  unposed real captures, \method recovers cleaner and more complete
  geometry than ReconViaGen.}
  \label{fig:co3d_qual}
\end{figure}

We quantitatively evaluate real-world generalization on CO3D: real
casual captures with cluttered backgrounds, spanning $20$ categories
with $2$ sequences each and $4$ unposed views per sequence. Since
CO3D provides only SfM point clouds as ground truth, predictions of
both methods are aligned to them with an identical Sim(3)+ICP
registration before evaluation. \method outperforms ReconViaGen on
both metrics (Table~\ref{tab:co3d}); Fig.~\ref{fig:co3d_qual} shows
qualitative comparisons.

\begin{table}[t]
\centering
\caption{\textbf{Quantitative real-world evaluation on CO3D.}
Real casual captures with cluttered backgrounds ($20$ categories
$\times$ $2$ sequences, $4$ unposed views each). Predictions of both
methods are aligned to the SfM ground-truth point clouds with an
identical Sim(3)~+~ICP procedure.}
\label{tab:co3d}
\vspace{-1em}
\small
\begin{tabular}{lcc}
\toprule
\textbf{Method} & \textbf{CD$\downarrow$} & \textbf{F-Score$\uparrow$} \\
\midrule
ReconViaGen & 0.037 & 0.80 \\
\textbf{\method (Ours)} & \textbf{0.029} & \textbf{0.84} \\
\bottomrule
\end{tabular}
\vspace{-0.5em}
\end{table}

\section{\texorpdfstring{Source-of-Gain Analysis with the Untouched Base Generator}{Source-of-Gain Analysis with the Untouched Base Generator}}
\label{sec:supp_ladder}

To disentangle the contribution of our design from the pretrained
generative backbone, we condition the \emph{pretrained}
Hunyuan3D-Omni---official weights, untouched---through its native
point-cloud interface and vary only the conditioning source:
(a)~off-the-shelf VGGT; (b)~our canonicalized VGGT (branch
repurposing + Sim(3) alignment); (c)~our full method with generator
fine-tuning. All four views are fused into the conditioning point
cloud, with one appearance view for the single-view image branch,
under the GSO protocol of Table~1 in the main paper.
Step (a)$\rightarrow$(b) isolates the reconstruction-side design and
(b)$\rightarrow$(c) the generator-side design; both yield clear gains
(Table~\ref{tab:ladder}), indicating that the improvement stems from
the proposed components rather than the backbone alone.

\begin{table}[t]
\centering
\caption{\textbf{Source-of-gain analysis on GSO.} The pretrained
Hunyuan3D-Omni (official weights) is conditioned through its native
point-cloud control; only the reconstruction source varies between
(a) and (b). Configuration (c) is our full method with generator
fine-tuning. (a)$\rightarrow$(b) isolates the reconstruction-side
design; (b)$\rightarrow$(c) isolates the generator-side design.}
\label{tab:ladder}
\vspace{-1em}
\resizebox{\linewidth}{!}{
\begin{tabular}{lcc}
\toprule
\textbf{Configuration} & \textbf{F-Score$\uparrow$} & \textbf{CD$\downarrow$} \\
\midrule
(a) Base generator + off-the-shelf VGGT & 0.155 & 0.153 \\
(b) Base generator + canonicalized VGGT (ours) & 0.567 & 0.038 \\
(c) \textbf{Full method (ours)} & \textbf{0.738} & \textbf{0.019} \\
\bottomrule
\end{tabular}
}
\vspace{-0.5em}
\end{table}

\section{\texorpdfstring{Backbone-Matched Comparison with ReconViaGen}{Backbone-Matched Comparison with ReconViaGen}}
\label{sec:supp_backbone_matched}

ReconViaGen builds on a different generative backbone (TRELLIS),
which confounds a direct comparison of the two conditioning
philosophies. We therefore re-instantiate its implicit feature
injection on the \emph{same} Hunyuan3D-Omni generator: learnable
query tokens cross-attend to frozen multi-layer VGGT features and
multi-view DINO tokens, and are injected into the DiT via
cross-attention; the geometry encoder, image tokens, training data,
schedule, and evaluation are identical to ours. On our training
validation set, explicit canonical-geometry conditioning outperforms
implicit feature injection on every metric
(Table~\ref{tab:backbone_matched}), indicating that the advantage
stems from the conditioning interface rather than the backbone.

\begin{table}[t]
\centering
\caption{\textbf{Backbone-matched comparison.} ReconViaGen's
implicit feature-injection is re-implemented on the same
Hunyuan3D-Omni backbone, with identical encoder, image tokens, data,
training budget, and evaluation. Both variants are trained for the
identical full schedule.}
\label{tab:backbone_matched}
\vspace{-1em}
\resizebox{\linewidth}{!}{
\begin{tabular}{lcccc}
\toprule
\textbf{Conditioning scheme} & \textbf{F-Score$\uparrow$} & \textbf{CD$\downarrow$} & \textbf{IoU$\uparrow$} & \textbf{Normal$\uparrow$} \\
\midrule
Implicit feature injection (ReconViaGen-style) & 0.399 & 0.0506 & 0.548 & 0.723 \\
\textbf{Explicit canonical geometry (ours)} & \textbf{0.771} & \textbf{0.0193} & \textbf{0.820} & \textbf{0.875} \\
\bottomrule
\end{tabular}
}
\vspace{-0.5em}
\end{table}

\section{\texorpdfstring{Joint Training vs. Decoupled Modular Training}{Joint Training vs. Decoupled Modular Training}}
\label{sec:supp_joint}

To quantify the disparate-learning-dynamics challenge, we train
three variants that share an identical backbone, data, and training
schedule, differing only in training mode: (a)~our decoupled design,
with the reconstruction module frozen during generator training;
(b)~joint training initialized from the same pretrained weights;
(c)~joint training from scratch. Joint training degrades every metric
even from identical initialization, and further from scratch
(Table~\ref{tab:joint_training}); with an equal budget, the
degradation is attributable to the coupling itself rather than to an
under-trained encoder.

\begin{table}[t]
\centering
\caption{\textbf{Joint training vs. decoupled modular training.}
All runs share the same backbone, data, and training schedule,
differing only in training mode.
Coupling the two objectives degrades every metric, even from
identical pretrained initialization.}
\label{tab:joint_training}
\vspace{-1em}
\resizebox{\linewidth}{!}{
\begin{tabular}{lccc}
\toprule
\textbf{Training mode} & \textbf{F-Score$\uparrow$} & \textbf{CD$\downarrow$} & \textbf{IoU$\uparrow$} \\
\midrule
\textbf{(a) Decoupled modular (ours)} & \textbf{0.70} & \textbf{0.0202} & \textbf{0.773} \\
(b) Joint, same pretrained init & 0.59 & 0.0229 & 0.737 \\
(c) Joint, from scratch & 0.54 & 0.0257 & 0.706 \\
\bottomrule
\end{tabular}
}
\vspace{-0.5em}
\end{table}

\section{\texorpdfstring{Generalization to a Different Reconstruction Backbone}{Generalization to a Different Reconstruction Backbone}}
\label{sec:supp_pi3}

\begin{figure}[H]
  \centering
    \includegraphics[width=\linewidth]{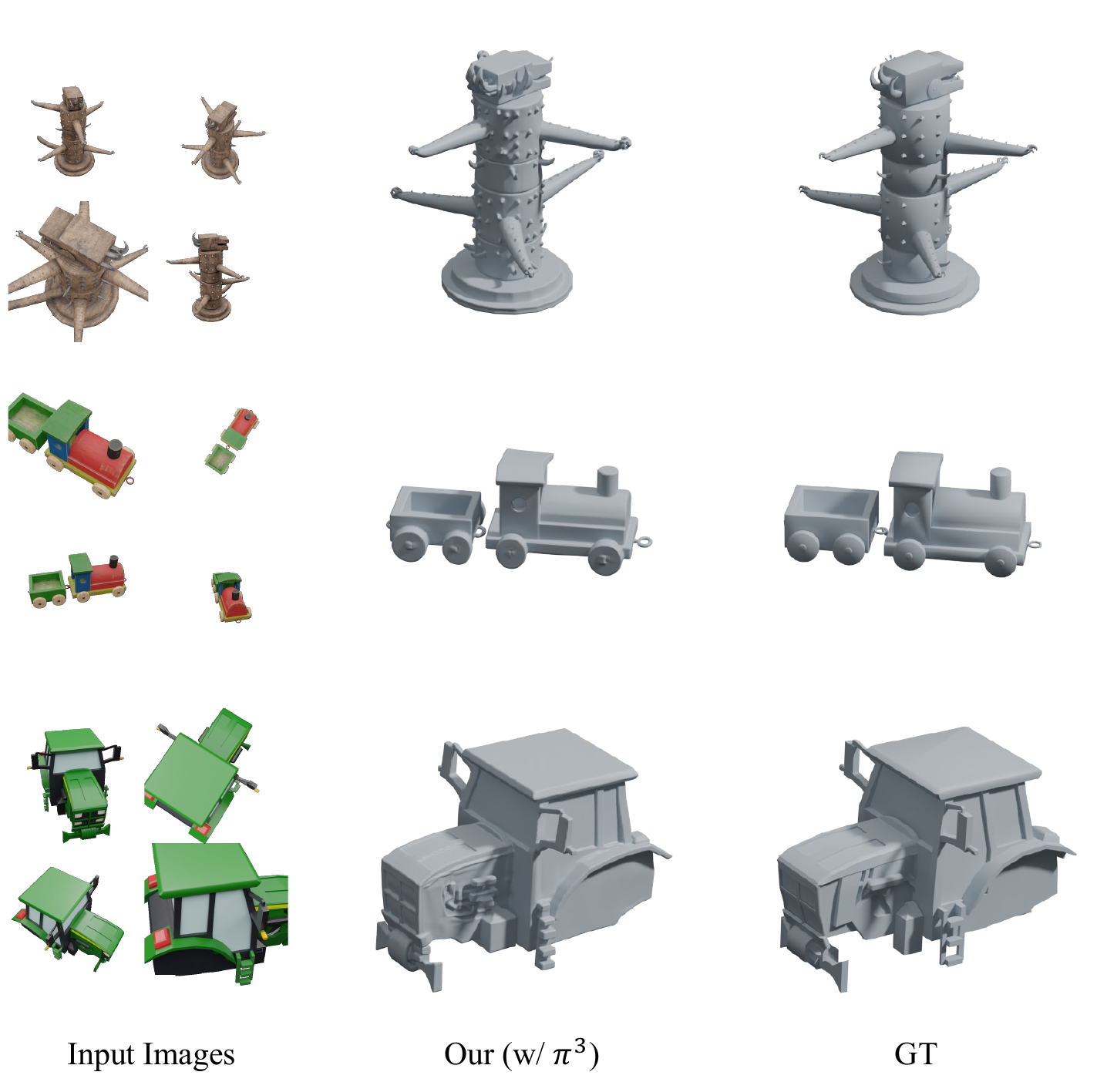}
  \caption{\textbf{Qualitative results with the $\pi^3$ backbone.}
  Replacing VGGT with $\pi^3$ yields accurate and complete meshes.}
  \label{fig:pi3_qual}
\end{figure}

Our framework treats the feed-forward reconstruction model as a
swappable module. To verify this, we replace VGGT with $\pi^3$
[Wang et~al.\ 2025], keeping the branch-repurposing recipe, training
data, and schedule identical. The $\pi^3$-based variant also attains
strong geometry (Table~\ref{tab:pi3_swap}; qualitative results in
Fig.~\ref{fig:pi3_qual}), indicating that the proposed
canonicalization is not specific to VGGT.

\begin{table}[t]
\centering
\caption{\textbf{Reconstruction backbone swap.} We replace VGGT
with $\pi^3$, keeping everything else identical; both variants are
trained under the same training schedule.}
\label{tab:pi3_swap}
\vspace{-1em}
\small
\begin{tabular}{lcccc}
\toprule
\textbf{Backbone} & \textbf{F-Score$\uparrow$} & \textbf{CD$\downarrow$} & \textbf{IoU$\uparrow$} & \textbf{Normal$\uparrow$} \\
\midrule
$\pi^3$ & 0.646 & 0.023 & 0.741 & 0.847 \\
VGGT & 0.762 & 0.020 & 0.816 & 0.873 \\
\bottomrule
\end{tabular}
\vspace{-0.5em}
\end{table}

\section{\texorpdfstring{Multi-Run Conditioning Ablation}{Multi-Run Conditioning Ablation}}
\label{sec:supp_multirun}

The conditioning ablation in the main paper (Table~5) uses a single
evaluation run over a $50$-object set. Since generation is stochastic
(diffusion sampling and randomized point sampling), we repeat the
evaluation with three inference runs under different random seeds on
an extended $100$-object set. Latent-augmented conditioning
consistently outperforms point-guided sampling, with gaps exceeding
one standard deviation (Table~\ref{tab:multirun}).

\begin{table}[t]
\centering
\caption{\textbf{Multi-run conditioning ablation.} Mean $\pm$
standard deviation over three inference runs with different random
seeds, evaluated on an extended $100$-object set.}
\label{tab:multirun}
\vspace{-1em}
\resizebox{\linewidth}{!}{
\begin{tabular}{lccc}
\toprule
\textbf{Strategy} & \textbf{CD$\downarrow$} & \textbf{F-Score$\uparrow$} & \textbf{IoU$\uparrow$} \\
\midrule
Point-Guided Sampling & 0.0207$\pm$0.0017 & 0.692$\pm$0.015 & 0.742$\pm$0.015 \\
\textbf{Latent-Augmented (ours)} & \textbf{0.0190$\pm$0.0015} & \textbf{0.720$\pm$0.010} & \textbf{0.760$\pm$0.016} \\
\bottomrule
\end{tabular}
}
\vspace{-0.5em}
\end{table}

\section{\texorpdfstring{Additional Implementation Details}{Additional Implementation Details}}
\label{sec:supp_impl}

\paragraph{Conditioning encoders.}
The view and camera projections in Eq.~(6) of the main paper are
implemented as lightweight single-layer projections. The VGGT
features $\mathbf{F}_i^{V}$ are formed by concatenating, along the
channel dimension, the outputs of four intermediate aggregator
blocks (blocks 4, 11, 17, and 23), each consisting of frame- and
global-attention features ($2\times1024$ channels), yielding 8192
channels in total; $\text{MLP}_{\text{view}}$ is a linear layer
($8192 \rightarrow 1024$) followed by RMSNorm and GELU. The camera
embedding uses the refined camera token of the final block (2048
channels); $\text{MLP}_{\text{cam}}$ is a linear layer
($2048 \rightarrow 1024$) followed by RMSNorm and GELU. The output
width of 1024 matches the conditioning width of the pretrained
Hunyuan3D-Omni encoder.

\paragraph{Point sampling.}
For the similarity alignment in Sec.~3.2 of the main paper, we set
$M=4096$: we first uniformly draw $4M=16384$ candidates from the
points that pass the reconstruction validity mask and confidence
filtering, and then apply farthest point sampling (with a random
start point) to obtain the $M$ spatially diverse correspondences;
evaluation fixes a global random seed for reproducibility. The same
$M=4096$ points serve both as correspondences for the Sim(3)
alignment and as the conditioning point cloud passed to the
generator.

\paragraph{Multi-view token layout.}
At the training resolution of $518\times518$ with patch size 14,
each input view contributes $K = 37^2 = 1369$ latent-augmented
tokens; the DINO class token and the VGGT camera/register prefix
tokens are dropped so that the two token grids align for the
element-wise addition in Eq.~(6). The $N$ per-view token sets are
concatenated along the sequence dimension, followed by the control
embedding $\beta$, which itself consists of $M=4096$ point-cloud
tokens (a Fourier positional encoding of the conditioning points,
linearly projected to width 1024) and $10$ learnable
modality-embedding tokens. The full conditioning sequence therefore
has length $NK + M + 10$; for the default $4$-view setting this
amounts to $4\times1369 + 4096 + 10 = 9582$ tokens. The base model's
single-view image tokens are \emph{replaced} by the $N$-view
latent-augmented tokens; the DiT cross-attends to exactly this
sequence, with no separate image-conditioning branch.

\paragraph{Inference cost breakdown.}
On a single NVIDIA RTX A6000 with four input views and $50$ denoising
steps, the full pipeline takes $\sim$40s per object at a peak memory
footprint of $\sim$16GB, versus $\sim$27s for the baseline
Hunyuan3D-Omni. The canonicalized reconstruction (including the
Sim(3) alignment) runs once per object; the remaining overhead comes
from the longer multi-view cross-attention context, which raises the
per-step diffusion time from $\sim$0.47s to $\sim$0.69s. Running the
two stages sequentially keeps peak memory bounded by the larger
stage.

\end{document}